\documentclass[10pt,twocolumn,letterpaper]{article}
\usepackage[T1]{fontenc}
\usepackage[margin=0.8in]{geometry}
\usepackage{times}
\usepackage{graphicx}
\usepackage[dvipsnames]{xcolor}
\usepackage{amssymb}
\usepackage{booktabs}
\usepackage[numbers,sort&compress]{natbib}
\usepackage{etoolbox}
\usepackage[format=plain,labelformat=simple,labelsep=period,font=small]{caption}
\usepackage[hyphens]{url}
\usepackage{algorithm}
\usepackage{algpseudocode}
\usepackage{mathtools}
\usepackage{tabularx}
\usepackage{multirow}
\usepackage{colortbl}
\usepackage{newfloat}
\usepackage{listings}
\usepackage{amsthm}
\usepackage{placeins}
\definecolor{kvsoftheader}{HTML}{EEF4FB}
\definecolor{kvsoftblue}{HTML}{BBD5EE}
\definecolor{kvsoftcyan}{HTML}{DDEBF7}
\definecolor{kvsoftgray}{HTML}{E7E6E6}

\DeclareCaptionStyle{ruled}{labelfont=normalfont,labelsep=colon,strut=off}
\floatstyle{ruled}
\newfloat{listing}{tb}{lst}{}
\floatname{listing}{Listing}

\definecolor{linkblue}{HTML}{2E75B6}
\usepackage[breaklinks,colorlinks,allcolors=linkblue]{hyperref}

\title{\bfseries LoopPrune: Recurrent Pruning for Looped Diffusion Transformers}
\date{}
\author{
Weizhong Huang\textsuperscript{1}\quad Huaijia Kou\textsuperscript{1}\quad Yan Zhang\textsuperscript{1}\quad Fei Chao\textsuperscript{1}\quad Xiawu Zheng\textsuperscript{1}\quad Liujuan Cao\textsuperscript{1}\thanks{Corresponding author.}\\
\textsuperscript{1}Key Laboratory of Multimedia Trusted Perception and Efficient Computing,\\
Ministry of Education of China, Xiamen University, 361005, P.R. China.
}
\newcommand{\method}{\textup{\textsc{LoopPrune}}}

\newcommand{\E}{\mathbb E}

\AtBeginEnvironment{thebibliography}{\small\setlength{\bibsep}{-.5pt}\setlength{\labelsep}{2pt}}

\hypersetup{pdftitle={LoopPrune: Recurrent Pruning for Looped Diffusion Transformers},pdfauthor={Weizhong Huang, Huaijia Kou, Yan Zhang, Fei Chao, Xiawu Zheng, Liujuan Cao},pdfsubject={arXiv preprint}}

\begin{document}

\raggedbottom

\maketitle

\setcounter{dbltopnumber}{2}
\setcounter{topnumber}{2}
\setcounter{bottomnumber}{1}
\setcounter{totalnumber}{4}
\renewcommand{\dbltopfraction}{0.92}
\renewcommand{\topfraction}{0.92}
\renewcommand{\bottomfraction}{0.92}
\renewcommand{\textfraction}{0.05}
\setlength{\textfloatsep}{3pt plus 1pt minus 1pt}
\setlength{\dbltextfloatsep}{3pt plus 1pt minus 1pt}
\setlength{\floatsep}{3pt plus 1pt minus 1pt}
\setlength{\dblfloatsep}{3pt plus 1pt minus 1pt}
\setlength{\intextsep}{3pt plus 1pt minus 1pt}
\captionsetup[figure]{skip=3pt}
\captionsetup[table]{skip=3pt}

\begin{abstract}
Looped diffusion Transformers reuse shared blocks to deepen computation and match larger models with fewer parameters, reducing storage and deployment burdens. Yet repeated attention and MLP computations remain costly. Pruning reduces storage and computation by removing redundant parameters or units. Training-free methods struggle to resolve recurrent contributions, while training-based methods lack explicit constraints on position-specific errors affecting later inputs and final outputs. We propose \method{}, a pruning and recovery framework addressing these requirements. Recurrent importance estimation combines shared-removal effects and position variation to select a shared structure; shared reconstruction pools recurrent input statistics to compensate weights. Loop-position recovery compares complete denoising predictions at identical student states and matches text guidance, updating all student parameters against a fixed dense teacher. Extensive experiments across four open-source models, six benchmarks, and three pruning paradigms demonstrate effectiveness. At 30\% structured parameter pruning, Looped-DiT B/16 retains \textbf{93.14\%} of dense six-benchmark average accuracy, exceeding the best baseline by \textbf{9.49\%}. Across PyTorch, vLLM-Omni, and SGLang Diffusion, pruned SDXSv3 reduces peak allocated GPU memory by \textbf{34.36--44.64\%} and accelerates denoising \textbf{1.46--1.74$\times$} over dense models.
\end{abstract}

\begin{figure}[tb]
\centering
\includegraphics[width=\linewidth,trim=0 4bp 0 0,clip]{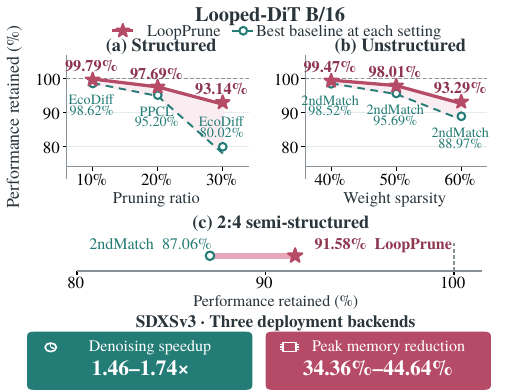}
\caption{\textbf{Performance and efficiency of \method{}.} B/16: retention of six-benchmark mean accuracy relative to Dense. SDXSv3: GPU memory savings and denoising speedups. Teal marks each setting's strongest baseline.}
\label{fig:teaser}
\end{figure}

\section{Introduction}
\label{sec:intro}

Diffusion Transformers (DiTs) improve generation by scaling capacity and depth~\citep{ddpm,latentdiffusion,dit,sd3}, increasing storage and deployment memory. Looped diffusion Transformers reuse blocks within each denoising step, matching larger DiTs with fewer parameters and deeper computation~\citep{loopeddit,sdxsv3,surjo}. Repeated attention and MLP operations remain costly, motivating further compression~\citep{huang2026commitkv}.

Network pruning~\citep{huang2025losa,huang2025atp,huang2025shapleymoe,zhang2026spt} removes less important parameters and structural units to reduce storage and computation~\citep{han2015learning,molchanov2019importance} while largely preserving diffusion generation quality~\citep{mosaicdiff,obsdiff,importanceobs}. Methods are training-free or training-based, depending on whether additional training is required.

\noindent\textbf{Training-free pruning.} Calibration statistics score importance or reconstruct weights without gradient training. Classical Wanda~\citep{wanda} combines weight and activation magnitudes, while SparseGPT and SlimGPT~\citep{sparsegpt,slimgpt} reconstruct weights. For diffusion, MosaicDiff~\citep{mosaicdiff} uses pretraining dynamics; OBS-Diff and Importance-aware OBS~\citep{obsdiff,importanceobs} use second-order reconstruction. These methods may overlook recurrent importance variation and error propagation, impairing selection and compensation.

\noindent\textbf{Training-based pruning.} Training learns masks or updates weights to restore quality. EcoDiff~\citep{ecodiff} optimizes structural masks; PPCL~\citep{ppcl} combines pruning, optimization, and distillation; MobileWan~\citep{mobilewan} optimizes head masks and weights. For unstructured pruning, 2ndMatch~\citep{secondmatch} uses second-order Jacobian matching. These objectives do not explicitly constrain position-specific pruning effects on later inputs and final outputs, limiting recovery from recurrent state shifts.

We propose \method{}, a pruning and recovery framework for looped diffusion Transformers. Recurrent importance estimation combines shared-removal effects and position variation to avoid underestimating canceling contributions and select a shared structure within budget. Under the fixed mask, shared projection reconstruction pools calibration statistics across positions to compensate retained weights for different recurrent inputs. Pruning and reconstruction initialize compression, but errors still alter later states. Loop-position recovery compares student and teacher stages at identical student states through the same teacher continuation. Matching complete conditional, unconditional, and guided predictions, together with states, updates all student parameters against a fixed teacher to accommodate recurrent state shifts.

We evaluate Looped-DiT B/16 and B/32, SDXSv3, and Surjo-Image-Preview on six benchmarks covering text alignment, reasoning, and spatial understanding under structured, semi-structured, and unstructured pruning. \method{} consistently achieves overall gains across models, benchmarks, and pruning paradigms. At 30\% structured parameter pruning, B/16 retains 93.14\% of dense six-benchmark average accuracy, exceeding the best baseline by 9.49\%. Across PyTorch, vLLM-Omni, and SGLang Diffusion, pruned SDXSv3 reduces peak allocated GPU memory by 34.36--44.64\% and accelerates denoising 1.46--1.74$\times$ over dense models, preserving generation accuracy while improving deployment efficiency.

\section{Related Work}
\label{sec:related}

\paragraph{Looped Diffusion Transformers.}
DiTs improve generation by scaling capacity and depth~\citep{dit,sd3}. Looped-DiT~\citep{loopeddit}, SDXSv3~\citep{sdxsv3}, and Surjo-Image-Preview~\citep{surjo} deepen computation through shared blocks with fewer parameters. Repeated attention and MLP operations still incur substantial memory and inference costs, which we reduce by pruning these blocks.

\paragraph{Network Pruning.}
Pruning removes less important parameters or structures~\citep{han2015learning,molchanov2019importance}. Training-free methods use statistics: Wanda~\citep{wanda} scores weight and activation magnitudes; SparseGPT~\citep{sparsegpt} and SlimGPT~\citep{slimgpt} reconstruct weights. For diffusion, MosaicDiff~\citep{mosaicdiff} uses pretraining dynamics, while OBS-Diff~\citep{obsdiff} and Importance-aware OBS~\citep{importanceobs} use second-order reconstruction. Training-based methods learn masks or update weights: EcoDiff~\citep{ecodiff} optimizes structural masks; PPCL~\citep{ppcl} combines pruning and distillation; MobileWan~\citep{mobilewan} optimizes head masks and weights; 2ndMatch~\citep{secondmatch} uses second-order Jacobian matching for sparse recovery. We address shared-unit reuse and pruning-error propagation in looped diffusion models.

\begin{figure*}[t]
\centering
\includegraphics[width=\textwidth,height=180pt]{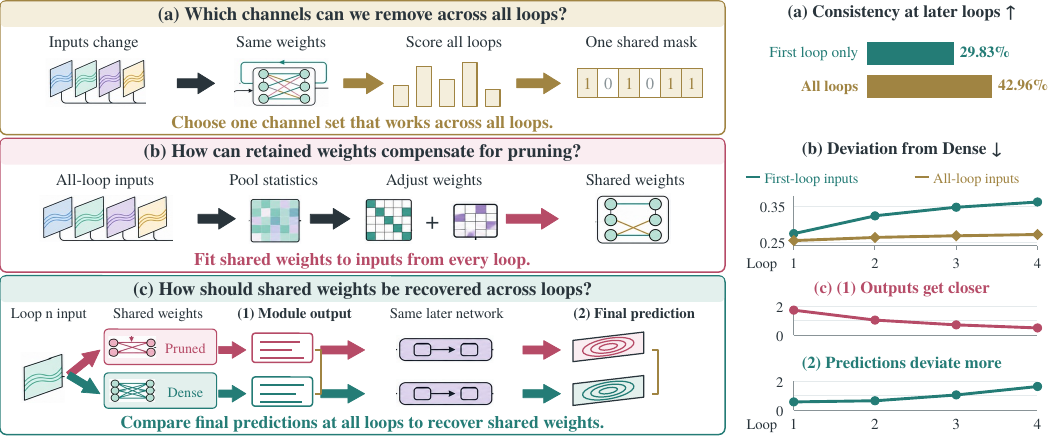}
\caption{\textbf{Why should pruning consider all loops?} (a) All-loop channel selection better agrees with later positions. (b) All-loop reconstruction reduces output deviations across positions. (c) Identical inputs enter pruned and dense modules at each position: compare outputs at (1) and final predictions after identical remaining computation at (2). Output deviation falls while prediction deviation rises, motivating all-loop prediction matching. Looped-DiT B/16, 30\% structured pruning; definitions and full analysis: Section~\ref{sec:observations} and Appendix~A.1.}

\label{fig:motivation}
\end{figure*}

\section{Preliminaries and Motivation}
\label{sec:observation}

\subsection{Preliminaries}
\label{sec:Problem_Formulation}

\paragraph{Looped diffusion Transformers.}
Looped diffusion models contain patch embeddings, a text encoder, recurrent Transformer blocks~\citep{attention}, and an output head. Noisy image/latent $\boldsymbol{x}_t$ is projected into patch tokens; with encoded text $\boldsymbol c$, these enter pre-loop $\mathcal A$ to form $\boldsymbol h^{(0)}$, before shared $\mathcal B$ and post-loop $\mathcal C$:
\begin{equation}
\boldsymbol{h}^{(0)}=\mathcal{A}(\boldsymbol{x}_t,\boldsymbol{c}),\quad
\boldsymbol{h}^{(n)}=\mathcal{B}\bigl(\boldsymbol{h}^{(n-1)}\bigr),\quad
\hat{\boldsymbol{x}}_t=\mathcal{C}\bigl(\boldsymbol{h}^{(n_{\ell})}\bigr),
\label{eq:looped}
\end{equation}
$\hat{\boldsymbol{x}}_t$ denotes the denoising prediction. Each denoising step calls $\mathcal B$ $n_{\ell}$ times, reusing its attention and MLP weights to update hidden states. Starting from noise, the sampler progressively updates the image or latent variables using the denoising prediction from Equation~\ref{eq:looped} and text guidance~\citep{cfg}; when denoising takes place in latent space, the final latent variables are passed through a VAE decoder to obtain a pixel-space image~\citep{latentdiffusion}.

\paragraph{Pruning and recovery.}
Under a budget, pruning $\mathcal P$ removes whole heads or MLP channels (structured)~\citep{michel2019sixteen}, zeros individual weights (unstructured)~\citep{wanda}, or enforces 2:4 masks (semi-structured)~\citep{sparsegpt}. For each target unit, a binary mask $m_j=1$ means retention and $m_j=0$ means pruning; all recurrent positions share the same pruning mask. Calibration statistics yield training-free reconstruction $\mathcal R_{\rm cal}$~\citep{obs,sparsegpt,slimgpt}; gradient recovery optimizes $\mathcal L_{\rm rec}$~\citep{ppcl,secondmatch}. Dense, pruned, reconstructed, and recovered weights follow:
\begin{equation}
\boldsymbol W
\xrightarrow{\mathcal P}\boldsymbol W_{\rm p}
\xrightarrow{\mathcal R_{\rm cal}}\widetilde{\boldsymbol W}_{\rm p}
\xrightarrow{\nabla\mathcal L_{\rm rec}}\boldsymbol W^{\star}.
\label{eq:pruning_recovery_pipeline}
\end{equation}

\newcommand{\obsBeforeAfterRho}{0.98}
\newcommand{\obsGateAfterRho}{0.27}
\newcommand{\obsCandidateFlips}{7}
\newcommand{\obsCandidatePairs}{223}
\newcommand{\obsDecisionChanges}{1}
\newcommand{\obsJointClipI}{0.795}
\newcommand{\obsJointClipIDelta}{+0.038}
\newcommand{\obsJointClipILow}{+0.019}
\newcommand{\obsJointClipIHigh}{+0.057}
\newcommand{\obsImageTradeoff}{Pixel MSE changes from 0.281 to 0.292. The output-only search control has CLIP-T 29.40, compared with 29.83 for the complete objective. Feature-space fidelity, pixel agreement, and text alignment therefore need separate checks.}

\subsection{Motivation}
\label{sec:recurrent_motivation}
Section~\ref{sec:Problem_Formulation} separates pruning, reconstruction, and gradient recovery. Shared parameters repeatedly process different inputs, with errors affecting later computation. We ask: \textbf{what special requirements does repeated use of shared modules place on pruning and recovery?} We examine channel selection, weight compensation, and prediction recovery to identify their dependence on complete recurrent computation.

Figure~\ref{fig:motivation} compares these stages on Looped-DiT B/16 at 30\% structured pruning across four recurrent positions, i.e., successive shared-module calls within a denoising step. These diagnostics identify the recurrent information each stage needs; Section~\ref{sec:observations} gives full measurements.

\paragraph{Which channels to prune?}
Single-loop importance may overlook a channel's role elsewhere. Figure~\ref{fig:motivation}a compares selection from the first loop alone versus all loops. Against channels selected separately at each later position, all-loop selection agrees better. We therefore \textbf{combine importance across all loops to choose one common set of retained channels}.

\paragraph{How to compensate retained weights?}
Single-loop reconstruction may not accommodate other positions' inputs. With identical retained channels, Figure~\ref{fig:motivation}b compares first-loop versus all-loop reconstruction. On independent inputs, all-loop reconstruction reduces output deviations at every position. We therefore \textbf{pool inputs from all loops to reconstruct shared weights}.

\paragraph{What should recovery compare?}
Shared weights receive different states across positions and influence subsequent computation. Figure~\ref{fig:motivation}c compares pruned and dense modules at identical inputs at four positions: first their outputs at (1), then predictions after identical remaining computation at (2). Across positions, output deviation falls while prediction deviation rises. Thus, output matching alone cannot reliably measure position-specific prediction effects. We \textbf{match final denoising predictions at all positions alongside intermediate features to recover one shared set of weights}.

These questions correspond to the algorithm designs for pruning, reconstruction, and recovery in Sections~\ref{sec:unrolled}--\ref{sec:loop_intervention}, respectively.

\section{Recurrent Pruning and Recovery}
\label{sec:method}
\begin{figure*}[t]
\centering
\includegraphics[width=\textwidth,height=170pt]{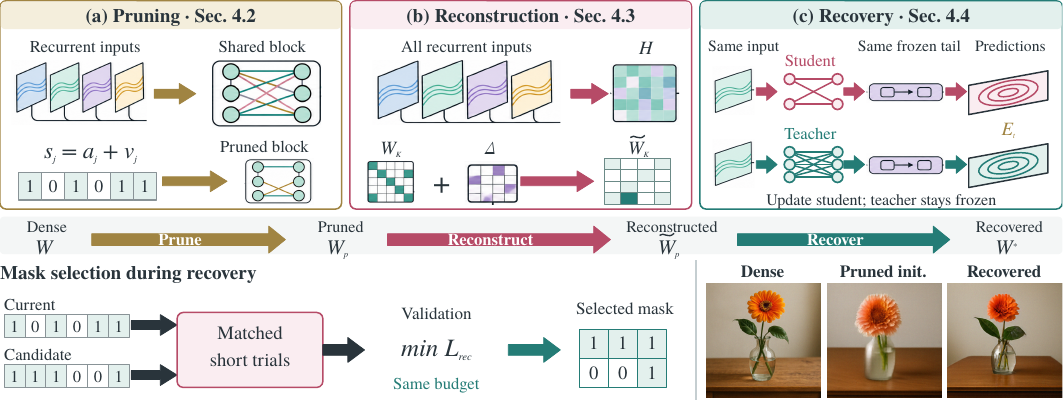}
\caption{\textbf{\method{} pruning--reconstruction--recovery pipeline.} (a) Score all positions and select a shared mask, obtaining $\boldsymbol W_{\rm p}$. (b) Pool recurrent inputs to compensate retained weights without training, obtaining $\widetilde{\boldsymbol W}_{\rm p}$. (c) Recover student $\boldsymbol W^\star$ by matching complete predictions and text guidance to a fixed teacher. Arrows connect stage outputs; matched trials select budget-preserving mask exchanges. Images share prompt and noise: recovery brings the vase, leaves, and layout closer to Dense.}
\label{fig:overview}
\end{figure*}

\subsection{Insight and Method Overview}
\label{sec:method_overview}

Our pipeline (Eq.~\eqref{eq:pruning_recovery_pipeline}, Figure~\ref{fig:overview}) addresses Section~\ref{sec:recurrent_motivation}'s questions through pruning, training-free reconstruction, and gradient recovery, transforming dense weights $\boldsymbol W$ into $\boldsymbol W_{\rm p}$, $\widetilde{\boldsymbol W}_{\rm p}$, and $\boldsymbol W^\star$.

Section~\ref{sec:unrolled} aggregates importance across positions to select a shared structure; Section~\ref{sec:semantic_reconstruction} fixes it and pools recurrent inputs to reconstruct weights. Section~\ref{sec:loop_intervention} recovers the reconstructed student against a fixed dense teacher through final-prediction, intermediate-feature, and text-guidance matching, with budget-preserving mask exchanges. Export retains the loop count and sampling procedure. Section~\ref{sec:observations} analyzes these designs.

\subsection{Recurrent-Aware Pruning}
\label{sec:unrolled}

Pruning $\boldsymbol W\xrightarrow{\mathcal P}\boldsymbol W_{\rm p}$ in Eq.~\eqref{eq:pruning_recovery_pipeline} removes the same units at every recurrent position, so we aggregate importance across positions. Units are attention heads or MLP channels for structured pruning, and individual weights for unstructured and semi-structured pruning.

Let $j$ index a unit and $n$ a recurrent position. Calibration freezes dense weights and assigns independent differentiable masks $m_{j,n}$: one retains the unit and zero prunes it; measurement uses all-one masks. Randomly adding or subtracting complete prediction components yields a scalar whose gradient $g_{j,n}$ with respect to $m_{j,n}$ describes its response to mask changes. One backward pass measures all positions without updating weights or masks.

Let $\E$ average over calibration inputs and random signs, and $\bar g_j$ denote the mean gradient across positions. Shared effect $a_j$ estimates removal at all positions; variation $v_j$ captures contributions canceled by gradient summation. Their sum is the base importance score $s_j$:
\begin{gather}
a_j=\E\!\left[\left(\sum_n g_{j,n}\right)^2\right],\quad
v_j=\E\!\left[\sum_n(g_{j,n}-\bar g_j)^2\right],\notag\\
s_j=a_j+v_j.\label{eq:recurrent_risk}
\end{gather}
\begin{figure}[!htbp]
\centering
\includegraphics[width=\columnwidth,trim=0 3bp 0 8bp,clip]{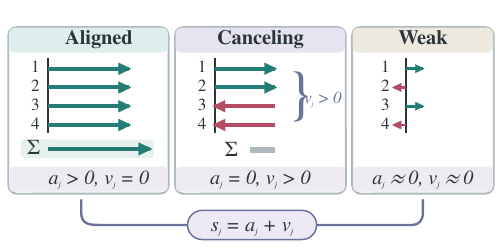}
\caption{\textbf{Why importance uses two terms.} Arrows show aligned, canceling, or weak recurrent gradients. Canceling gradients can deviate strongly from their mean (large $v_j$), whereas weak gradients have small deviations and $v_j$.}
\label{fig:recurrent_importance_schematic}
\end{figure}

In Figure~\ref{fig:recurrent_importance_schematic}, aligned strong gradients yield large $a_j$; opposing strong gradients yield near-zero $a_j$ and mean but large deviations and $v_j$; weak gradients yield small values for both. Thus, $v_j$ distinguishes cancellation from weak effects.
We prune low-$s_j$ units to the target rate, enforcing per-row retained counts for unstructured pruning and keeping the two highest-scoring weights per consecutive four-weight row group for 2:4. All positions share the mask.

\subsection{Training-Free Shared Reconstruction}
\label{sec:semantic_reconstruction}

Training-free reconstruction $\boldsymbol W_{\rm p}\xrightarrow{\mathcal R_{\rm cal}}\widetilde{\boldsymbol W}_{\rm p}$ implements Eq.~\eqref{eq:pruning_recovery_pipeline} of Section~\ref{sec:Problem_Formulation}. We fix the shared mask selected by $s_j$ in Section~\ref{sec:unrolled} and adjust retained weights using calibration inputs to compensate for removed contributions.

Shared weights receive different inputs across positions, so we pool their calibration statistics. Let $\boldsymbol W$ be the original projection weights, $\boldsymbol x$ the input features, and $\mathcal K,\mathcal R$ the mask's retained and removed feature indices. The pooled input-statistics matrix is
\begin{equation}
\boldsymbol H=\E_{\omega}[\boldsymbol x\boldsymbol x^\top],
\label{eq:reconstruction_statistics}
\end{equation}
where $\E_{\omega}$ is weighted averaging over calibration inputs. $\boldsymbol H_{\mathcal K\mathcal K}$ records statistics within retained features, and $\boldsymbol H_{\mathcal R\mathcal K}$ between removed and retained features.

\begin{figure}[!htbp]
\centering
\includegraphics[width=\columnwidth,trim=0 3bp 0 0,clip]{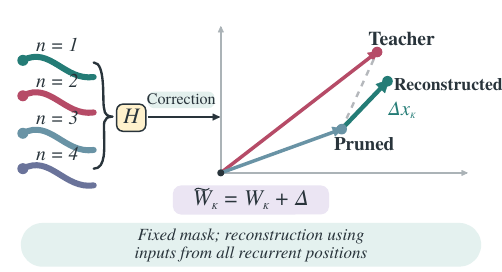}
\caption{\textbf{Shared reconstruction.} Pooled recurrent inputs estimate $\boldsymbol H$. Under the fixed mask, correction $\Delta\boldsymbol x_{\mathcal K}$ moves the pruned projection output toward the teacher output.}
\label{fig:shared_reconstruction_schematic}
\end{figure}

The original output sums retained and removed contributions, $\boldsymbol W_{\mathcal K}\boldsymbol x_{\mathcal K}$ and $\boldsymbol W_{\mathcal R}\boldsymbol x_{\mathcal R}$. We add correction $\Delta$ to retained weights so $\Delta\boldsymbol x_{\mathcal K}$ approximates the removed output (Figure~\ref{fig:shared_reconstruction_schematic}). Damping $\lambda>0$ limits correction magnitude and stabilizes least squares~\citep{golub2013matrix}:
\begin{equation}
\min_{\Delta}\;\E_{\omega}
\left\|\boldsymbol W_{\mathcal R}\boldsymbol x_{\mathcal R}-\Delta\boldsymbol x_{\mathcal K}\right\|^2
+\lambda\|\Delta\|_F^2.
\label{eq:reconstruction_objective}
\end{equation}
This ridge-regression objective balances output compensation and correction magnitude~\citep{hoerl1970ridge}.

Let $\boldsymbol I$ be the identity matrix. Differentiating the objective with respect to $\Delta$ and setting the derivative to zero gives the linear system $\Delta(\boldsymbol H_{\mathcal K\mathcal K}+\lambda\boldsymbol I)=\boldsymbol W_{\mathcal R}\boldsymbol H_{\mathcal R\mathcal K}$. The compensated retained weights $\widetilde{\boldsymbol W}_{\mathcal K}=\boldsymbol W_{\mathcal K}+\Delta$ are therefore
\begin{equation}
\widetilde{\boldsymbol{W}}_{\mathcal{K}}=\boldsymbol{W}_{\mathcal{K}}+\boldsymbol{W}_{\mathcal{R}}\boldsymbol{H}_{\mathcal{R}\mathcal{K}}(\boldsymbol{H}_{\mathcal{K}\mathcal{K}}+\lambda \boldsymbol{I})^{-1},
\label{eq:recurrent_repair}
\end{equation}
For unstructured and 2:4 pruning, apply Eq.~\eqref{eq:recurrent_repair} to each output weight row $i$ with its own retained/removed column indices $\mathcal K_i,\mathcal R_i$ and corresponding statistics submatrices. Updating only retained weights preserves pruned zeros and the 2:4 mask.

Cholesky factorization~\citep{golub2013matrix} solves this system without gradient training. Applying each projection's compensated weights $\widetilde{\boldsymbol W}_{\mathcal K}$ to $\boldsymbol W_{\rm p}$ gives $\widetilde{\boldsymbol W}_{\rm p}$ for Section~\ref{sec:loop_intervention}; other parameters and the shared pruning structure remain unchanged.

\subsection{Gradient-Based Recurrent Recovery}
\label{sec:loop_intervention}
\label{sec:mask_refinement}
\label{sec:short_recovery_method}

Recovery implements $\widetilde{\boldsymbol W}_{\rm p}\xrightarrow{\nabla\mathcal L_{\rm rec}}\boldsymbol W^{\star}$ in Eq.~\eqref{eq:pruning_recovery_pipeline}. Reconstruction (Section~\ref{sec:semantic_reconstruction}) initializes student $\mathcal S$; dense teacher $\mathcal D$ stays fixed. Since pruning alters later recurrent states, we recover complete predictions and text guidance.

\paragraph{Comparing complete denoising predictions.}
Matching stage outputs alone does not explicitly constrain their effect on subsequent computation. We feed the same student features into student and teacher stages, then pass their outputs through identical remaining teacher computation to compare complete current-step predictions (Figure~\ref{fig:overview}c).

\begin{figure}[!htbp]
\centering
\includegraphics[width=\columnwidth,trim=0 2bp 0 8bp,clip]{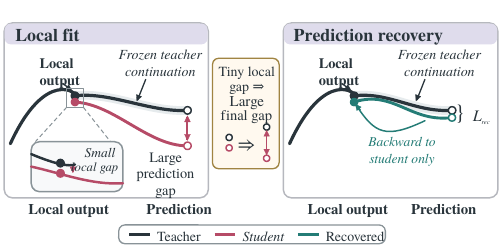}
\caption{\textbf{Why compare complete predictions?} Black/rose/teal: teacher/student/recovered student. Left: a small local gap can grow after identical teacher continuation. Right: recovery targets the complete current-step prediction gap; the backward arrow updates the student through the fixed teacher.}
\label{fig:prediction_recovery_schematic}
\end{figure}

Figure~\ref{fig:prediction_recovery_schematic} shows teacher (black), student (rose), and recovered student (teal). The left inset shows local gaps growing through teacher continuation; the right bracket targets teal predictions toward black, with gradients returning through the fixed teacher to the student.

Let $r$ denote the recurrent position, $\boldsymbol h^{\mathcal S}_{r-1}$ the student features before it, $\mathcal B_{\mathcal S}$ and $\mathcal B_{\mathcal D}$ the student and teacher shared stages, and $\mathcal T_{\mathcal D,r}$ the remaining teacher network and output head. In this comparison, $\mathcal G$ fixes input features, blocking gradients to earlier computation while allowing updates to the selected student stage. The complete predictions are
\begin{equation}
\boldsymbol{a}_r=\mathcal{T}_{\mathcal{D},r}(\mathcal{B}_{\mathcal{S}}(\mathcal{G}[\boldsymbol{h}^{\mathcal{S}}_{r-1}])),\quad
\boldsymbol{b}_r=\mathcal{T}_{\mathcal{D},r}(\mathcal{B}_{\mathcal{D}}(\mathcal{G}[\boldsymbol{h}^{\mathcal{S}}_{r-1}])).
\label{eq:loop_intervention}
\end{equation}

\begin{algorithm}[t]
\caption{\method{} pruning and recovery overview.}
\label{alg:loopprune_overview}
\normalsize
\begin{algorithmic}[1]
\Require Teacher, budget, calibration/training datasets
\State Estimate $s_j$ across all positions (Sec.~\ref{sec:unrolled})
\State Select a shared mask within budget (Sec.~\ref{sec:unrolled})
\State Pool inputs; reconstruct retained weights (Sec.~\ref{sec:semantic_reconstruction})
\State Initialize student; freeze teacher (Sec.~\ref{sec:loop_intervention})
\State Refine the mask with matched trials (Sec.~\ref{sec:loop_intervention})
\For{each recovery step}
  \State Select recurrent position $r$ (Sec.~\ref{sec:loop_intervention})
  \State Compare predictions at identical inputs (Sec.~\ref{sec:loop_intervention})
  \State Match conditional, unconditional, and guided predictions; compute $\mathcal L_{\rm rec}$ (Sec.~\ref{sec:loop_intervention})
  \State Update all student parameters: $\nabla\mathcal L_{\rm rec}$ (Sec.~\ref{sec:loop_intervention})
\EndFor
\State Export; retain loop count and sampler (Sec.~\ref{sec:loop_intervention})
\end{algorithmic}
\end{algorithm}

The discrepancy between student-stage prediction $\boldsymbol a_r$ and teacher-stage prediction $\boldsymbol b_r$, called propagated error, measures replacement effects at a common input. Gradients pass through the fixed teacher continuation to the student stage. Training covers all recurrent positions.

\paragraph{Simultaneously recovering text guidance.}
Even small conditional and unconditional prediction errors can distort their difference, which determines text guidance. We match both predictions and the guided prediction constructed from their difference. The teacher's conditional--unconditional difference emphasizes text-sensitive regions. Their three errors form $\mathcal E_t$, where $t$ is denoising time.

\paragraph{Recovery objective and weight updates.}
The term $\mathcal E_t(\mathcal S,\mathcal D)$ compares complete student and teacher predictions~\citep{distill} at the same noisy input to constrain overall pruning error; $\mathcal E_t(\boldsymbol a_r,\boldsymbol b_r)$ compares Eq.~\eqref{eq:loop_intervention} predictions to constrain the selected position's effect. Auxiliary loss $\mathcal L_h$ matches shared-stage output features~\citep{fitnets}. The default recovery loss is
\begin{equation}
\mathcal{L}_{\rm rec}=\mathcal{E}_t(\mathcal{S},\mathcal{D})+\beta_r\,\mathcal{E}_t(\boldsymbol{a}_r,\boldsymbol{b}_r)+\beta_h\mathcal{L}_h.
\label{eq:joint_loss}
\end{equation}
Here $\beta_r$ and $\beta_h$ weight the position prediction error and state-matching loss, respectively. Backpropagation updates all student parameters while teacher and external-encoder parameters remain fixed. Algorithm~\ref{alg:loopprune_overview} summarizes the pruning and recovery procedure.

\begin{table*}[t!]
\centering
\footnotesize
\setlength{\tabcolsep}{2pt}
\renewcommand{\arraystretch}{0.85}
\setlength{\aboverulesep}{0.25ex}
\setlength{\belowrulesep}{0.25ex}
\caption{\textbf{Representative results across models and pruning formats.} Average is the unweighted mean of six benchmarks; bold marks the best pruning score within a configuration.}
\label{tab:summary_results}
\begin{tabularx}{\textwidth}{@{}>{\raggedright\arraybackslash}p{0.30\textwidth}*{7}{>{\centering\arraybackslash}X}@{}}
\toprule
\textbf{Method} & \textbf{GenEval} & \textbf{DPG} & \textbf{TIIF} & \textbf{PRISM} & \textbf{CoRe} & \textbf{Spatial} & \textbf{Average} \\
\midrule
\multicolumn{8}{c}{\cellcolor{kvsoftheader}\textbf{Looped-DiT B/16 ($n_{\ell}=4$) --- 30\% structured parameter pruning}}\\
\midrule
\rowcolor{kvsoftgray}Dense & 86.06 & 86.98 & 76.14 & 66.49 & 53.67 & 64.61 & 72.32 \\
Wanda-sp~(ICLR'24) & 51.67 & 66.51 & 58.91 & 28.74 & 30.47 & 50.07 & 47.73 \\
SlimGPT~(NeurIPS'24) & 62.57 & 75.90 & 62.38 & 36.93 & 37.00 & 60.07 & 55.81 \\
MosaicDiff~(ICCV'25) & 29.76 & 64.68 & 47.20 & 31.06 & 32.55 & 50.04 & 42.55 \\
OBS-Diff~(ICLR'26) & 61.72 & 74.06 & 59.93 & 40.10 & 35.99 & 57.85 & 54.94 \\
Importance-aware OBS~(NeurIPS'26) & 53.97 & 70.02 & 57.28 & 35.32 & 34.09 & 54.73 & 50.90 \\
EcoDiff~(ICLR'26) & 61.55 & 77.68 & 67.72 & 36.61 & 42.42 & 61.26 & 57.87 \\
PPCL~(CVPR'26) & 60.76 & 75.04 & 63.29 & 35.79 & 37.97 & 59.73 & 55.43 \\
MobileWan~(NeurIPS'26) & 47.46 & 70.15 & 59.66 & 31.00 & 37.40 & 56.92 & 50.43 \\
\rowcolor{kvsoftcyan}\textbf{\method{}} & \textbf{84.06} & \textbf{83.44} & \textbf{71.81} & \textbf{50.13} & \textbf{50.09} & \textbf{64.63} & \textbf{67.36} \\
\midrule
\multicolumn{8}{c}{\cellcolor{kvsoftheader}\textbf{Looped-DiT B/16 ($n_{\ell}=4$) --- 2:4 semi-structured sparsity}}\\
\midrule
\rowcolor{kvsoftgray}Dense & 86.06 & 86.98 & 76.14 & 66.49 & 53.67 & 64.61 & 72.32 \\
Wanda~(ICLR'24) & 7.65 & 43.51 & 37.92 & 15.97 & 19.85 & 39.15 & 27.34 \\
SparseGPT~(ICML'23) & 43.42 & 65.04 & 53.54 & 29.24 & 27.90 & 50.99 & 45.02 \\
OBS-Diff~(ICLR'26) & 46.09 & 66.41 & 53.25 & 30.42 & 28.15 & 51.66 & 46.00 \\
Importance-aware OBS~(NeurIPS'26) & 56.48 & 70.10 & 59.13 & 33.29 & 31.21 & 54.18 & 50.73 \\
2ndMatch~(CVPR'26) & 76.80 & 81.03 & 67.75 & 47.21 & 42.99 & 61.96 & 62.96 \\
\rowcolor{kvsoftcyan}\textbf{\method{}} & \textbf{82.95} & \textbf{83.19} & \textbf{71.40} & \textbf{48.04} & \textbf{48.05} & \textbf{63.77} & \textbf{66.23} \\
\midrule
\multicolumn{8}{c}{\cellcolor{kvsoftheader}\textbf{Looped-DiT B/16 ($n_{\ell}=4$) --- 50\% unstructured sparsity}}\\
\midrule
\rowcolor{kvsoftgray}Dense & 86.06 & 86.98 & 76.14 & 66.49 & 53.67 & 64.61 & 72.32 \\
Wanda~(ICLR'24) & 78.32 & 83.73 & 70.82 & 49.79 & 48.93 & 63.89 & 65.91 \\
SparseGPT~(ICML'23) & 77.56 & 82.79 & 70.70 & 54.89 & 47.42 & 63.65 & 66.17 \\
OBS-Diff~(ICLR'26) & 77.63 & 83.25 & 70.93 & 54.39 & 48.27 & 63.42 & 66.32 \\
Importance-aware OBS~(NeurIPS'26) & 79.46 & 83.17 & 71.51 & 54.75 & 48.39 & 64.32 & 66.93 \\
2ndMatch~(CVPR'26) & 82.82 & 85.15 & 73.05 & 58.43 & 50.84 & 64.94 & 69.20 \\
\rowcolor{kvsoftcyan}\textbf{\method{}} & \textbf{86.25} & \textbf{85.92} & \textbf{75.41} & \textbf{60.25} & \textbf{51.98} & \textbf{65.46} & \textbf{70.88} \\
\midrule
\multicolumn{8}{c}{\cellcolor{kvsoftheader}\textbf{SDXSv3 ($n_{\ell}=2$) --- 50\% unstructured sparsity}}\\
\midrule
\rowcolor{kvsoftgray}Dense & 26.35 & 63.57 & 50.09 & 40.36 & 29.98 & 51.71 & 43.67 \\
Wanda~(ICLR'24) & 14.96 & 51.19 & 44.15 & 29.50 & 21.93 & 40.83 & 33.76 \\
SparseGPT~(ICML'23) & 24.26 & 61.12 & 49.00 & 37.76 & 28.63 & 51.41 & 42.03 \\
OBS-Diff~(ICLR'26) & 23.72 & 62.06 & 48.84 & 37.97 & 28.34 & 52.24 & 42.20 \\
Importance-aware OBS~(NeurIPS'26) & 23.86 & 61.81 & 49.13 & 38.17 & 28.12 & 50.61 & 41.95 \\
2ndMatch~(CVPR'26) & 24.75 & 63.62 & \textbf{49.99} & 37.31 & \textbf{31.40} & 52.90 & 43.33 \\
\rowcolor{kvsoftcyan}\textbf{\method{}} & \textbf{26.30} & \textbf{63.97} & 49.89 & \textbf{39.41} & 30.23 & \textbf{53.00} & \textbf{43.80} \\
\midrule
\multicolumn{8}{c}{\cellcolor{kvsoftheader}\textbf{Surjo-Image-Preview ($n_{\ell}=2$) --- 2:4 semi-structured sparsity}}\\
\midrule
\rowcolor{kvsoftgray}Dense & 32.92 & 73.77 & 52.33 & 46.09 & 34.23 & 52.93 & 48.71 \\
Wanda~(ICLR'24) & 12.74 & 59.89 & 41.45 & 28.38 & 23.21 & 41.63 & 34.55 \\
SparseGPT~(ICML'23) & 19.09 & 62.83 & 43.45 & 31.11 & 22.99 & 44.80 & 37.38 \\
OBS-Diff~(ICLR'26) & 20.16 & 62.26 & 43.05 & 31.69 & 23.83 & 45.23 & 37.70 \\
Importance-aware OBS~(NeurIPS'26) & 19.17 & 63.27 & 44.63 & 31.20 & 24.91 & 45.42 & 38.10 \\
2ndMatch~(CVPR'26) & 19.42 & 61.49 & 44.35 & 28.93 & 24.28 & 43.86 & 37.06 \\
\rowcolor{kvsoftcyan}\textbf{\method{}} & \textbf{22.16} & \textbf{66.58} & \textbf{47.53} & \textbf{35.79} & \textbf{27.32} & \textbf{48.98} & \textbf{41.39} \\
\bottomrule
\end{tabularx}
\end{table*}

\begin{figure*}[t]
\centering
\includegraphics[width=\textwidth,trim=0 10bp 0 0,clip]{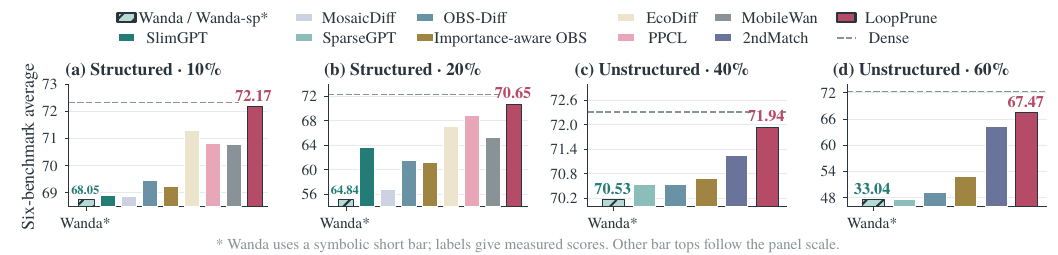}
\caption{\textbf{Additional pruning budgets on B/16.} (a,b) 10\%/20\% structured parameter pruning; (c,d) 40\%/60\% unstructured sparsity. Rose denotes \method{}; gray dashed lines mark dense. Wanda* uses a symbolic short bar labeled with its measured score; other bar tops follow the panel scale.}
\label{fig:budget_results}
\end{figure*}

\paragraph{Shared-mask refinement.}
Recovery changes weights and which units are worth retaining, so we adjust the initial mask to adapt both to recurrent computation. Each round starts with four weight updates, using gradients of $\mathcal L_{\rm rec}$ with respect to the shared mask to reassess importance. A candidate replaces lower-scoring retained units with higher-scoring masked units under the same budget and sparsity constraints: MLP channels for structured pruning and weight positions for sparse pruning. Full-shaped weights remain available so masked units can be reactivated.

In Figure~\ref{fig:overview}, current and candidate masks start from identical weights and optimizer states and receive four updates on the same batches. Comparing $\mathcal L_{\rm rec}$ on the same independent validation data selects the lower-loss mask and its updated weights. Each mask is shared across all positions. After four selection rounds, we fix the shared mask for the remaining recovery. Student weights continue to adapt to recurrent states, while all positions retain the same selected units. This preserves the pruning budget and sparsity constraints; export retains the original loop count and sampling procedure. Further settings appear in Appendix~D.1.

\makeatletter
\setlength{\@fptop}{0pt}
\makeatother

\section{Experiments}
\label{sec:experiments}

\subsection{Experimental Setup}
\label{sec:experimental_setup}

\paragraph{Models.}
We evaluate four-loop Looped-DiT B/16 and B/32~\citep{loopeddit}, and two-loop SDXSv3~\citep{sdxsv3} and Surjo-Image-Preview~\citep{surjo}. The 11 settings comprise B/16 at 10\%/20\%/30\% structured, 40\%/50\%/60\% unstructured, and 2:4 semi-structured pruning; B/32 at 2:4 and 50\% unstructured; SDXSv3 at 50\% unstructured; and Surjo at 2:4. Architectures: Appendix~B.

\paragraph{Baselines.}
Structured baselines are Wanda-sp, SlimGPT, MosaicDiff, OBS-Diff, Importance-aware OBS, EcoDiff, PPCL, and MobileWan~\citep{wanda,slimgpt,mosaicdiff,obsdiff,importanceobs,ecodiff,ppcl,mobilewan}. Semi-structured and unstructured baselines are Wanda, SparseGPT, OBS-Diff, Importance-aware OBS, and 2ndMatch~\citep{wanda,sparsegpt,obsdiff,importanceobs,secondmatch}. Settings appear in Appendix~C.1.

\paragraph{Benchmarks.}
We assess alignment on GenEval~\citep{geneval}, DPG-Bench~\citep{ella}, and TIIF-Bench short~\citep{tiif}, and reasoning and spatial understanding on PRISM-Bench~\citep{prism}, T2I-CoReBench~\citep{corebench}, and SpatialGenEval~\citep{spatialgeneval}. Average is their unweighted mean. Coverage, prompts, judges, and generation settings: Appendix~C.2.

\paragraph{Implementation Details.}
We use 256 COCO captions~\citep{coco} for calibration and recovery, with eight H20 GPUs and 208 recovery updates by default. Defaults are $\lambda=1\%$ of the mean diagonal of $\boldsymbol H_{\mathcal K\mathcal K}$ and $\beta_r=\beta_h=0.25$. Further settings: Appendix~D.1.

\subsection{Main Results}
\label{sec:main_results}
\label{sec:completed_b16_results}
\newcommand{\resBStructTenAvg}{72.17}
\newcommand{\resBStructTenBest}{71.32}
\newcommand{\resBStructTenGain}{0.85}
\newcommand{\resBStructTenDeficit}{-0.85}
\newcommand{\resBStructTenGenEval}{86.21}
\newcommand{\resBStructTwentyAvg}{70.65}
\newcommand{\resBStructTwentyBest}{68.85}
\newcommand{\resBStructTwentyGain}{1.80}
\newcommand{\resBStructTwentyDeficit}{-1.80}
\newcommand{\resBStructTwentyGenEval}{87.51}
\newcommand{\resBStructThirtyAvg}{67.36}
\newcommand{\resBStructThirtyBest}{57.87}
\newcommand{\resBStructThirtyGain}{9.49}
\newcommand{\resBStructThirtyDeficit}{-9.49}
\newcommand{\resBStructThirtyGenEval}{84.06}
\newcommand{\resBUnstructFortyAvg}{71.94}
\newcommand{\resBUnstructFortyBest}{71.25}
\newcommand{\resBUnstructFortyGain}{0.69}
\newcommand{\resBUnstructFortyDeficit}{-0.69}
\newcommand{\resBUnstructFortyGenEval}{86.77}
\newcommand{\resBUnstructFiftyAvg}{70.88}
\newcommand{\resBUnstructFiftyBest}{69.20}
\newcommand{\resBUnstructFiftyGain}{1.68}
\newcommand{\resBUnstructFiftyDeficit}{-1.68}
\newcommand{\resBUnstructFiftyGenEval}{86.25}
\newcommand{\resBUnstructSixtyAvg}{67.47}
\newcommand{\resBUnstructSixtyBest}{64.34}
\newcommand{\resBUnstructSixtyGain}{3.13}
\newcommand{\resBUnstructSixtyDeficit}{-3.13}
\newcommand{\resBUnstructSixtyGenEval}{82.71}
\newcommand{\resBSmallUnstructAvg}{65.90}
\newcommand{\resBSmallUnstructBest}{65.86}
\newcommand{\resBSmallUnstructGain}{0.04}
\newcommand{\resBSmallUnstructDeficit}{-0.04}
\newcommand{\resBSmallUnstructGenEval}{81.42}
\newcommand{\resSdxUnstructAvg}{43.80}
\newcommand{\resSdxUnstructBest}{43.33}
\newcommand{\resSdxUnstructGain}{0.47}
\newcommand{\resSdxUnstructDeficit}{-0.47}
\newcommand{\resSdxUnstructGenEval}{26.30}
\newcommand{\resBTwoFourAvg}{66.23}
\newcommand{\resBTwoFourBest}{62.96}
\newcommand{\resBTwoFourGain}{3.27}
\newcommand{\resBTwoFourDeficit}{-3.27}
\newcommand{\resBTwoFourGenEval}{82.95}
\newcommand{\resBSmallTwoFourAvg}{63.25}
\newcommand{\resBSmallTwoFourBest}{62.74}
\newcommand{\resBSmallTwoFourGain}{0.51}
\newcommand{\resBSmallTwoFourDeficit}{-0.51}
\newcommand{\resBSmallTwoFourGenEval}{79.44}
\newcommand{\resSurjoTwoFourAvg}{41.39}
\newcommand{\resSurjoTwoFourBest}{38.10}
\newcommand{\resSurjoTwoFourGain}{3.29}
\newcommand{\resSurjoTwoFourDeficit}{-3.29}
\newcommand{\resSurjoTwoFourGenEval}{22.16}
\newcommand{\resBDenseGenEval}{86.06}
\newcommand{\resBDenseAvg}{72.32}
\newcommand{\resBSmallDenseGenEval}{84.49}
\newcommand{\resBSmallDenseAvg}{67.30}
\newcommand{\resSdxDenseGenEval}{26.35}
\newcommand{\resSdxDenseAvg}{43.67}
\newcommand{\resSurjoDenseGenEval}{32.92}
\newcommand{\resSurjoDenseAvg}{48.71}

\paragraph{Comparison with State-of-the-Art Methods.}
\method{} leads pruning methods in six-benchmark average across Table~\ref{tab:summary_results}'s five settings, spanning three architectures and pruning formats. B/16 scores \resBStructThirtyAvg{}/\resBTwoFourAvg{}/\resBUnstructFiftyAvg{} at 30\% structured/2:4 semi-structured/50\% unstructured pruning, with absolute gains of \resBStructThirtyGain{}\%/\resBTwoFourGain{}\%/\resBUnstructFiftyGain{}\% over the strongest baselines (EcoDiff for structured pruning; 2ndMatch otherwise). It also leads every benchmark among pruning methods in these three B/16 settings.

\paragraph{Results at Different Pruning Ratios.}
\label{sec:different_pruning_ratios}
\method{} leads all pruning baselines at four additional B/16 budgets (Figure~\ref{fig:budget_results}). At 10\%/20\% structured pruning, \resBStructTenAvg{}/\resBStructTwentyAvg{} averages yield absolute gains of \resBStructTenGain{}\%/\resBStructTwentyGain{}\% over the strongest baselines; at 40\%/60\% unstructured pruning, \resBUnstructFortyAvg{}/\resBUnstructSixtyAvg{} yield \resBUnstructFortyGain{}\%/\resBUnstructSixtyGain{}\%. With Table~\ref{tab:summary_results}, these results show consistent advantages increasing at higher pruning rates.

\begin{figure}[t]
\centering
\includegraphics[width=\linewidth,height=104pt,trim=0 16bp 0 0,clip]{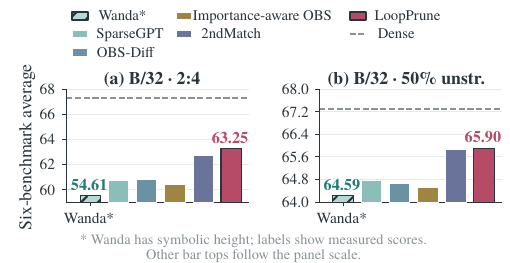}
\caption{\textbf{Results on Looped-DiT B/32.} (a) 2:4 and (b) 50\% unstructured sparsity. Rose denotes \method{}; dashed lines mark dense. Wanda* has symbolic height with its measured score labeled; other bar tops follow the panel scale.}
\label{fig:sparse_results}

\end{figure}

\paragraph{Results on Broader Model Architectures.}
\label{sec:models_pruning_paradigms}
B/32 scores \resBSmallTwoFourAvg{}/\resBSmallUnstructAvg{} at 2:4 semi-structured/50\% unstructured pruning (Figure~\ref{fig:sparse_results}), exceeding the strongest baseline, 2ndMatch, by \resBSmallTwoFourGain{}\%/\resBSmallUnstructGain{}\%. Table~\ref{tab:summary_results}'s SDXSv3 (50\% unstructured) and Surjo (2:4) score \resSdxUnstructAvg{}/\resSurjoTwoFourAvg{}, exceeding 2ndMatch/Importance-aware OBS by \resSdxUnstructGain{}\%/\resSurjoTwoFourGain{}\%. These absolute gains extend the B/16 results across Looped-DiT variants, two additional shared-block architectures, model sizes, and pruning formats.

\newcounter{looppruneHyperparameterFigure}
\setcounter{looppruneHyperparameterFigure}{\numexpr\value{figure}+2\relax}
\begingroup
\edef\looppruneBeforeAnalysisFigure{\number\value{figure}}
\setcounter{figure}{\value{looppruneHyperparameterFigure}}
\begin{figure*}[!t]
\centering
\includegraphics[width=\textwidth,height=108pt]{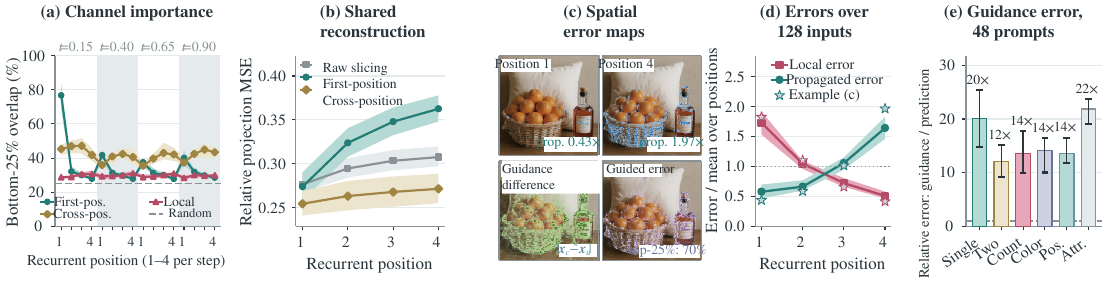}
\caption{\textbf{Recurrent pruning analysis.} (a) Overlap with each position's least-important 25\% channels; dashed: random. (b) Relative projection MSE, identical retained channels. (c) Positions 1/4 propagated errors (top, shared scale); teacher guidance difference/student guided-prediction error (bottom). (d) Median local/propagated errors, normalized by each input's four-position mean; stars: (c). Local error compares module outputs, whereas propagated error compares complete predictions after the same remaining teacher computation. (e) Median relative-error ratio: guidance difference/individual predictions. COCO~\citep{coco}/GenEval~\citep{geneval} protocols: Appendices~A.1,~A.2.}
\label{fig:observation}
\end{figure*}

\setcounter{figure}{\looppruneBeforeAnalysisFigure}
\endgroup

\subsection{Efficiency Analyses}
\label{sec:efficiency}
\label{sec:efficiency_analyses}

\begin{figure}[H]
\centering
\includegraphics[width=\linewidth,height=155pt]{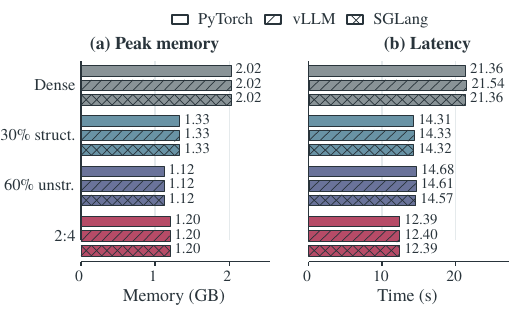}
\caption{\textbf{Measured SDXSv3 inference efficiency (lower is better).} Left: peak allocated memory (GB); right: batch latency (s). Hatches distinguish backends.}
\label{fig:efficiency_results}

\end{figure}

We benchmark SDXSv3 on one H20 with PyTorch, vLLM-Omni, and SGLang Diffusion (Figure~\ref{fig:efficiency_results}). The 30\% structured, 60\% unstructured, and 2:4 semi-structured models use compact matrices, custom sparse CUDA kernels, and cuSPARSELt, respectively. They reduce peak allocated memory by 34.36\%, 44.64\%, and 40.71\%, with denoising speedups of 1.49--1.50$\times$, 1.46--1.47$\times$, and 1.72--1.74$\times$ over each backend's displayed Dense latency. These savings hold across pruning formats and backends. Settings and optimization costs: Appendix~E.1.

\setcounter{figure}{\numexpr\value{looppruneHyperparameterFigure}+1\relax}
\subsection{Ablation Studies}
\label{sec:ablation_studies}

\paragraph{Effectiveness of Components.}
Table~\ref{tab:ablation_main} tests five ablations of Sections~\ref{sec:unrolled}--\ref{sec:loop_intervention}. Full averages 72.54 on GenEval, DPG-Bench, and PRISM. Removing position variation or shared reconstruction lowers this to 70.02 or 66.57. Removing loop-position prediction matching ($\beta_r=0$) or state matching ($\beta_h=0$) yields 70.32 or 70.17. Skipping gradient recovery gives 63.43, the largest drop. All components help; reconstruction and recovery separate initial weight compensation from gradient-based gains. Details: Appendix~E.2.

\begin{table}[t]
\centering
\caption{\textbf{Component ablations on B/16 at 30\% structured pruning.} Each variant removes one component from Full. Avg.\ (3) averages GenEval, DPG, and PRISM; bold marks column maxima.}
\label{tab:ablation_main}
\label{tab:ablation_mechanisms}
\setlength{\tabcolsep}{2pt}
\renewcommand{\arraystretch}{0.96}
\footnotesize
\begin{tabularx}{\linewidth}{@{}Xrrrr@{}}
\toprule
Variant & GenEval & DPG & PRISM & Avg. (3)\\
\midrule
\rowcolor{kvsoftcyan}\textbf{Full \method{}} & \textbf{84.06} & \textbf{83.44} & \textbf{50.13} & \textbf{72.54}\\
w/o position variation ($v_j=0$) & 80.51 & 81.17 & 48.39 & 70.02\\
w/o shared reconstruction (Sec.~\ref{sec:semantic_reconstruction}) & 75.05 & 80.31 & 44.36 & 66.57\\
w/o loop-position loss ($\beta_r=0$) & 80.90 & 82.05 & 48.01 & 70.32\\
w/o state matching ($\beta_h=0$) & 81.11 & 81.88 & 47.51 & 70.17\\
w/o gradient recovery (Sec.~\ref{sec:loop_intervention}) & 67.73 & 78.44 & 44.11 & 63.43\\
\bottomrule
\end{tabularx}
\end{table}

\paragraph{Hyperparameter Sensitivity.}
\label{sec:main_hyperparameter_sensitivity}
Figure~\ref{fig:ablation_hyperparameters} varies one coefficient across six non-default settings on B/16 at 30\% structured pruning. GenEval/DPG/PRISM Avg.\ (3) is 69.84--70.81 for reconstruction damping $\lambda$, 70.49--70.57 for loop-position weight $\beta_r$, and 70.08--70.17 for state-matching weight $\beta_h$. The defaults in Section~\ref{sec:experimental_setup} score 72.54, the highest tested. Details: Appendix~E.3.

\begingroup
\edef\looppruneRestoreFigure{\number\value{figure}}
\setcounter{figure}{\numexpr\value{looppruneHyperparameterFigure}-1\relax}
\begin{figure}[!t]
\centering
\includegraphics[width=\linewidth,height=85pt,trim=0 14bp 0 1bp,clip]{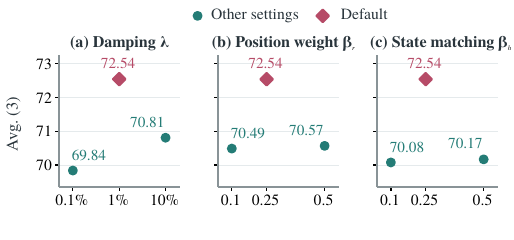}
\caption{\textbf{Hyperparameter sensitivity: B/16, 30\% structured pruning.} GenEval/DPG/PRISM mean versus one coefficient per panel; $\lambda$ is relative to $\boldsymbol H_{\mathcal K\mathcal K}$'s diagonal mean. Teal circles: other settings; rose diamonds: defaults.}
\label{fig:ablation_hyperparameters}
\end{figure}

\setcounter{figure}{\looppruneRestoreFigure}
\endgroup

\subsection{Further Analysis}
\label{sec:observations}

We revisit Section~\ref{sec:recurrent_motivation}'s three design questions on B/16 at 30\% structured pruning (Figure~\ref{fig:observation}). Importance and prediction diagnostics use 32 COCO captions at four denoising times (128 inputs), with weights fixed.

\paragraph{Cross-position importance.}
We select the lowest-scoring 25\% channels using first-position prediction effects, all-position local errors, or all-position prediction effects; each position's prediction-based selection is the reference. At later positions, first-position and local-error selections overlap this reference by 28--32\% and about 30\%, near the 25\% random baseline. Cross-position selection reaches 35--47\% (Figure~\ref{fig:observation}a). Importance varies across positions, and local errors may miss prediction effects, supporting a shared mask estimated across all positions (Section~\ref{sec:unrolled}).

\paragraph{Shared reconstruction.}
With identical retained channels, we fit on 32 COCO captions and evaluate projection errors against dense outputs on 32 independent captions, without gradient training. Relative projection MSE is 0.2546--0.2716 for all-position reconstruction, versus 0.2766--0.3077 for raw slicing and 0.2740--0.3627 for first-position fitting (Figure~\ref{fig:observation}b). First-position fitting deteriorates later; all-position fitting remains best, showing the benefit of compensating shared weights for all recurrent inputs (Section~\ref{sec:semantic_reconstruction}).

\paragraph{Prediction and text-guidance errors.}
Pruned and dense modules receive identical student states. We compare their outputs locally and complete predictions after the same teacher continuation. From positions 1 to 4, normalized median local error falls from 1.73 to 0.51, but propagated error rises from 0.58 to 1.64 (Figure~\ref{fig:observation}d). Figure~\ref{fig:observation}c illustrates larger prediction deviations and localized guidance errors. On 48 GenEval prompts spanning six categories, the conditional--unconditional difference, which determines text guidance, has 12$\times$--22$\times$ the relative error of the individual predictions (Figure~\ref{fig:observation}e). The dense teacher defines both relative-error reference values. Small local or individual-branch errors therefore do not ensure accurate predictions or guidance, supporting joint prediction, guidance, and feature matching (Section~\ref{sec:loop_intervention}).

\section{Conclusion}
\label{sec:conclusion}
\method{} combines recurrent-aware pruning, training-free shared reconstruction, and gradient-based recovery for looped diffusion Transformers. Experiments demonstrate overall advantages over existing pruning methods, preserving generation quality while reducing GPU memory use and accelerating denoising.

\FloatBarrier

\section*{AI Assistance Disclosure}
\phantomsection
\label{sec:ai_assistance}
We used OpenAI Codex to assist with writing, translation, experimental design, implementation, result analysis, and figure preparation. The authors take full responsibility for all content and conclusions.\par

\clearpage
\appendix
\section*{Appendix}
\pdfbookmark[0]{Appendix}{appendix_overview}
\raggedbottom
\setcounter{dbltopnumber}{3}
\setcounter{topnumber}{3}
\setcounter{bottomnumber}{2}
\setcounter{totalnumber}{4}
\renewcommand{\dbltopfraction}{0.93}
\renewcommand{\topfraction}{0.85}
\renewcommand{\bottomfraction}{0.50}
\renewcommand{\textfraction}{0.06}
\renewcommand{\floatpagefraction}{0.76}
\renewcommand{\dblfloatpagefraction}{0.76}
\setlength{\textfloatsep}{8pt plus 1pt minus 1pt}
\setlength{\dbltextfloatsep}{9pt plus 1pt minus 1pt}
\setlength{\floatsep}{10pt plus 2pt minus 2pt}
\setlength{\dblfloatsep}{10pt plus 2pt minus 2pt}
\setlength{\intextsep}{8pt plus 1pt minus 1pt}
\captionsetup{font=small,skip=4pt}
\captionsetup[figure]{skip=4pt}
\captionsetup[table]{skip=4pt}
\makeatletter
\setlength{\@fptop}{0pt}
\setlength{\@dblfptop}{0pt}
\setlength{\@fpsep}{12pt plus 2pt minus 2pt}
\setlength{\@dblfpsep}{12pt plus 2pt minus 2pt}
\setlength{\@fpbot}{0pt plus 1fil}
\setlength{\@dblfpbot}{0pt plus 1fil}
\makeatother
\newcommand{\appendixfigurewidth}{.92\textwidth}

\section{More Analysis}
\label{app:more_analysis}
\label{sec:measurement}

\subsection{Measurements of Figure~\ref{fig:observation}}
\label{sec:obs_v2_protocol}

The reconstruction protocol for Figures~\ref{fig:motivation}(b) and~\ref{fig:observation}(b) is described in Appendix~\ref{app:current_transfer}.

\paragraph{Channel importance (a).}
We use the frozen Looped-DiT B/16 EMA teacher and its 30\% structured-pruned initializer. Panels (a,c,d) use 32 held-out COCO captions at denoising steps 15, 40, 65, and 90 (128 inputs). For panel (a), we rank channels by mask gradients using the first position's complete-prediction error, all positions' local errors, or all positions' complete-prediction errors. We sum the scores over the sampled steps and select the lowest-scoring 25\% of channels in each shared MLP. The reference at each position uses that position's complete-prediction error, and we average overlaps across MLPs and captions.

\paragraph{Prediction errors (c,d).}
The pruned and dense modules take the same student state. Local error averages the image and text outputs' relative squared errors. Propagated error compares the complete predictions after the same teacher continuation (Eqs.~\eqref{eq:loop_intervention} and~\eqref{eq:paired_distance}). Panel (c) uses a common scale for positions 1 and 4. Its top-25\% label gives the share of guided error energy in the quarter of pixels with the largest dense guidance differences. Panel (d) divides each input's errors by their mean across the four positions before taking the median.

\paragraph{Text guidance (e).}
We evaluate eight GenEval prompts per task at steps 40 and 65 (96 inputs across six tasks). For each input, we divide the relative error of the conditional--unconditional prediction difference by the combined relative error of the two predictions, using the dense teacher as the reference. Each bar shows the median of these ratios for one task.
\subsection{Cross-Loop Transfer of Projection Reconstruction}
\label{app:current_transfer}
For Figure~\ref{fig:observation}b, we fix the retained channels and compare direct slicing, reconstruction from the first loop position, and reconstruction from all four positions. We fit the retained weights of four shared MLP projections on 32 COCO captions without gradient training and evaluate on 32 separate held-out captions. At each loop position, we measure the relative mean squared error against the dense projection outputs.
\FloatBarrier

\section{Detailed Model Architectures}
\label{app:model_setup}
\label{sec:Details_of_Models}

Table~\ref{tab:architecture_summary} summarizes the four denoisers.

\begin{table}[H]
\centering
\footnotesize
\setlength{\tabcolsep}{2pt}
\caption{\textbf{Architecture of the evaluated denoisers.} Model names link to their checkpoints.}
\label{tab:architecture_summary}
\begin{tabular*}{\linewidth}{@{\extracolsep{\fill}}llrrrr@{}}
\toprule
Model & \shortstack[l]{Input\\space} & Width & Heads & Blocks & Loops \\
\midrule
\href{https://huggingface.co/sensenova/Looped-DiT-B16}{Looped-DiT B/16}~\citep{loopeddit} & Pixel & 768 & 12 & $(6,5,6)$ & 4 \\
\href{https://huggingface.co/sensenova/Looped-DiT-B32}{Looped-DiT B/32}~\citep{loopeddit} & Pixel & 768 & 12 & $(6,5,6)$ & 4 \\
\href{https://huggingface.co/recoilme/sdxsv3}{SDXSv3}~\citep{sdxsv3} & Latent & 2{,}560 & 20 & $(2,12,2)$ & 2 \\
\href{https://huggingface.co/SurjoLabs/Surjo-Image-Preview}{Surjo-Image-Preview}~\citep{surjo} & Latent & 512 & 8 & $(7,4,7)$ & 2 \\
\bottomrule
\end{tabular*}
\end{table}

B/16 has 236.0M prunable parameters out of 258.1M. SDXSv3 uses loop-specific embeddings and RMSNorm. Surjo uses self- and cross-attention in shared blocks. Table~\ref{tab:flops} counts two FLOPs per multiply-add and counts the savings of a shared unit over all $n_{\ell}$ executions.

\begin{table}[H]
\centering
\captionsetup{skip=2pt}
\footnotesize
\setlength{\tabcolsep}{3pt}
\caption{\textbf{Analytic FLOPs of Looped-DiT per network evaluation.} Per-component FLOPs are for one execution in one MMDiT block; a head spans both streams.}
\label{tab:flops}
\resizebox{\linewidth}{!}{
\begin{tabular}{@{}lcc@{}}
\toprule
 & \textbf{B/16} & \textbf{B/32} \\
\midrule
Tokens (image + text) & 1{,}024 + 256 & 256 + 256 \\
Denoiser, 4 loops (GFLOPs) & 747.8 & 264.5 \\
Shared stage share & 61.9\% & 60.9\% \\
One block (GFLOPs) & 23.15 & 8.05 \\
\quad heads / image MLP / text MLP & 47.8 / 41.7 / 10.4\% & 40.0 / 30.0 / 30.0\% \\
One head (GFLOPs) & 0.923 & 0.268 \\
One image / text channel (MFLOPs) & 4.72 / 1.18 & 1.18 / 1.18 \\
\bottomrule
\end{tabular}}
\end{table}
\section{Baselines and Evaluation Settings}
\label{sec:Details_of_Models_Baselines_Benchmarks}

\subsection{Baseline Implementations}
\label{sec:Details_of_Baselines}

\paragraph{Training-free baselines.}
Wanda-sp~\citep{flap,wanda} pools calibration statistics across loops and guidance branches. SlimGPT~\citep{slimgpt} recalibrates pruned blocks. We match MosaicDiff~\citep{mosaicdiff} by step-weighted active parameters. OBS-Diff~\citep{obsdiff} combines rankings across timesteps, and Importance-aware OBS~\citep{importanceobs} weights tokens by the guidance difference. Wanda and SparseGPT~\citep{wanda,sparsegpt} use matched target matrices and sparsity constraints.

\paragraph{Training-based baselines.}
EcoDiff~\citep{ecodiff,hardconcrete}, PPCL~\citep{ppcl}, MobileWan~\citep{mobilewan}, and 2ndMatch~\citep{secondmatch} use matched pruning budgets and the same calibration and diagnostic data. 2ndMatch trains on a fixed OBS-Diff mask.

\subsection{Benchmark Settings and Judges}
\label{sec:Details_of_Benchmarks}

All methods use the same prompts, judges, and scoring coverage criteria. Table~\ref{tab:benchmark_details} lists the six benchmarks. SpatialGenEval uses one judge rollout per image. We average the six scores with equal weight.

\begin{table}[H]
\centering
\footnotesize
\setlength{\tabcolsep}{2pt}
\caption{\textbf{Benchmark evaluation settings.}}
\label{tab:benchmark_details}
\begin{tabular*}{\linewidth}{@{\extracolsep{\fill}}lcl@{}}
\toprule
\textbf{Benchmark} & \shortstack{\textbf{Images}\\\textbf{per prompt}} & \textbf{Judge} \\
\midrule
GenEval & 4 & Mask2Former~\citep{mask2former} + CLIP \\
DPG-Bench & 1 & mPLUG-large VQA~\citep{mplug} \\
TIIF-Bench (short) & 1 & Qwen3.8-27B~\citep{qwen38_27b} \\
PRISM-Bench (en) & 1 & Qwen2.5-VL-72B-Instruct~\citep{qwen25vl} \\
T2I-CoReBench & 1 & Qwen3-VL-32B-Thinking~\citep{qwen3vl} \\
SpatialGenEval & 1 & Qwen2.5-VL-72B-Instruct~\citep{qwen25vl} \\
\bottomrule
\end{tabular*}
\end{table}

\paragraph{Evaluation Coverage.}
In the order listed in Table~\ref{tab:benchmark_details}, the evaluated splits contain 553, 1,065, 2,538, 700, 1,080, and 1,230 prompts, respectively. We check that every prompt has the expected number of generated images. For scoring, we check for 2,212 image-level records on GenEval, 1,065 prompt-level scores on DPG-Bench, and 1,230 image-level records on SpatialGenEval. For TIIF-Bench, PRISM-Bench, and T2I-CoReBench, we account for every expected scoring item, including failed judgments, and check valid scoring coverage separately.

\paragraph{Generation.}
Looped-DiT, SDXSv3, and Surjo use 100, 28, and 50 sampling steps at guidance scales 6.0, 4.5, and 3.0, with output resolutions of $512\times512$, $1024\times1024$, and $256\times256$, respectively.
\section{Properties and Implementation}
\label{app:method_details}

\subsection{Optimization and Reproducibility Details}
\label{sec:short_recovery_details}

\paragraph{Recurrent-aware pruning.}
In Eq.~\eqref{eq:recurrent_risk}, the calibration gradient $g_{j,n}$ comes from a random sign projection of the complete prediction. Let $\boldsymbol p_t\in\mathbb R^d$ concatenate the dense model's conditional, unconditional, and normalized guided predictions with the weighting in Eq.~\eqref{eq:paired_distance}, where $d$ is the total number of prediction components after concatenation. Draw $\boldsymbol\xi\in\{-1,1\}^d$ with each component independently and uniformly sampled from $\{-1,1\}$, and compute $g_{j,n}=\partial\langle\boldsymbol\xi,\boldsymbol p_t\rangle/\partial m_{j,n}$ with all masks set to one. Budget accounting counts each shared parameter once and includes FLOPs from every invocation.

\paragraph{Training-free shared reconstruction.}
Equation~\eqref{eq:reconstruction_statistics} averages input outer products across recurrent positions. The weights $\omega$ emphasize valid conditional tokens. Pooling across positions allows one retained projection to compensate for the removed features across all its calls. Damping makes $\boldsymbol H_{\mathcal K\mathcal K}+\lambda\boldsymbol I$ positive definite, so Eq.~\eqref{eq:recurrent_repair} gives a unique correction. For weight-sparse masks, each output row uses the retained and removed indices of that row in the same pooled statistics.

\paragraph{Gradient-based recurrent recovery.}
The two predictions in Eq.~\eqref{eq:loop_intervention} use the same student input and teacher continuation, isolating the effect of replacing one shared stage. Equation~\eqref{eq:joint_loss} uses the same distance $\mathcal E_t$ for full-model and position-specific prediction matching. For prediction pairs $\boldsymbol a=(\boldsymbol a_c,\boldsymbol a_u)$ and $\boldsymbol b=(\boldsymbol b_c,\boldsymbol b_u)$, define the errors $\boldsymbol e_c=(\boldsymbol a_c-\boldsymbol b_c)/q_t$, $\boldsymbol e_u=(\boldsymbol a_u-\boldsymbol b_u)/q_t$, and $\boldsymbol e_g=[\boldsymbol e_u+s(\boldsymbol e_c-\boldsymbol e_u)]/s$, where $s$ is the guidance scale. For clean-image prediction, $q_t$ is the native noise scale with a positive lower bound. For velocity prediction, $q_t=1$.
\begin{equation}
\mathcal{E}_t(\boldsymbol{a},\boldsymbol{b})=
\frac{\langle \boldsymbol{w}_t(\boldsymbol{e}_c^2+\boldsymbol{e}_u^2+\boldsymbol{e}_g^2)/3\rangle}{\langle\boldsymbol{w}_t\rangle}.
\label{eq:paired_distance}
\end{equation}
Here $\langle\cdot\rangle$ averages prediction entries, and $\boldsymbol w_t=1+2\boldsymbol v_t$. We get $\boldsymbol v_t$ by averaging the absolute teacher guidance difference across channels and applying per-image min--max normalization, so regions more sensitive to text receive higher weight.

For state matching, we normalize the output difference by the teacher's feature energy in each native stream $u\in\mathcal U$, with a small positive floor $\epsilon$. In Looped-DiT, the streams are image and text.
\begin{equation}
\mathcal{L}_h=\frac{1}{|\mathcal{U}|}\sum_{u\in\mathcal{U}}
\frac{\langle(\boldsymbol{h}^{\mathcal{S}}_{r,u}-\boldsymbol{h}^{\mathcal{D}}_{r,u})^2\rangle}
{\max(\langle(\boldsymbol{h}^{\mathcal{D}}_{r,u})^2\rangle,\epsilon)}.
\label{eq:hidden_matching}
\end{equation}

\paragraph{Shared-mask refinement.}
We compare candidates on held-out captions. Each of four rounds uses four updates to propose an exchange and four updates per mask in the matched trial. Counting both trials and 160 fixed-mask updates gives the default total of 208 updates. Before exporting weight-sparse models, we reapply the shared mask to the recovered weights.
\section{More Results}
\label{app:more_results}
\subsection{Optimization Cost and Efficiency Protocol}
\label{app:efficiency_details}

\paragraph{Optimization cost.}
Table~\ref{tab:optimization_costs} reports costs with 256 fitting captions and batch size 32. Timing includes calibration probes, mask search, and recovery where applicable. Cached initializer construction for \method{} and OBS-Diff is excluded.

\begin{table*}[!tp]
\centering
\footnotesize
\setlength{\tabcolsep}{2.5pt}
\caption{\textbf{Measured optimization costs.} Each entry gives GPU-minutes / elapsed seconds / number of H20 GPUs.}
\label{tab:optimization_costs}
\begin{tabularx}{\textwidth}{@{}ll*{4}{>{\centering\arraybackslash}X}@{}}
\toprule
\multicolumn{6}{@{}l}{\textbf{(a) Structured pruning}} \\
\addlinespace[1pt]
Model & Pruning & \method{} & EcoDiff & PPCL & MobileWan \\
\midrule
Looped-DiT B/16 & 10\% structured & 76.94 / 577.1 / 8 & 203.49 / 1526.2 / 8 & 56.00 / 420.0 / 8 & 58.26 / 437.0 / 8 \\
Looped-DiT B/16 & 20\% structured & 33.32 / 249.9 / 8 & 194.63 / 1459.7 / 8 & 58.48 / 438.6 / 8 & 56.87 / 426.5 / 8 \\
Looped-DiT B/16 & 30\% structured & 33.16 / 248.7 / 8 & 195.51 / 1466.3 / 8 & 51.10 / 383.3 / 8 & 57.16 / 428.7 / 8 \\
\bottomrule
\end{tabularx}
\par\vspace{4pt}
\begin{tabularx}{\textwidth}{@{}ll*{2}{>{\centering\arraybackslash}X}@{}}
\toprule
\multicolumn{4}{@{}l}{\textbf{(b) Semi-structured and unstructured pruning}} \\
\addlinespace[1pt]
Model & Pruning & \method{} & 2ndMatch \\
\midrule
Looped-DiT B/16 & 2:4 & 38.04 / 285.3 / 8 & 100.26 / 1503.9 / 4 \\
Looped-DiT B/16 & 40\% unstructured & 33.59 / 251.9 / 8 & 93.42 / 1401.3 / 4 \\
Looped-DiT B/16 & 50\% unstructured & 39.04 / 292.8 / 8 & 104.27 / 782.0 / 8 \\
Looped-DiT B/16 & 60\% unstructured & 34.27 / 257.0 / 8 & 97.00 / 1455.0 / 4 \\
Looped-DiT B/32 & 2:4 & 22.55 / 169.2 / 8 & 105.14 / 1577.1 / 4 \\
Looped-DiT B/32 & 50\% unstructured & 22.66 / 170.0 / 8 & 105.53 / 1582.9 / 4 \\
SDXSv3 & 50\% unstructured & 182.19 / 1366.4 / 8 & 180.88 / 2713.1 / 4 \\
Surjo & 2:4 & 23.17 / 173.8 / 8 & 97.16 / 728.7 / 8 \\
\bottomrule
\end{tabularx}
\end{table*}

\paragraph{SDXSv3 inference.}
Figure~\ref{fig:efficiency_results} measures denoising latency and peak allocated GPU memory over seven post-warmup runs on one H20. All three backends use the same inputs and weights for each setting. Runs use batch size 16, $768\times768$ resolution, BF16, two recurrent positions, and 28 Euler steps at guidance scale 4.5.
\subsection{Component Ablations}
\label{app:ablation_design}

Table~\ref{tab:ablation_main} uses B/16 at 30\% structured pruning. All variants share the fitting and validation data and generation noise. Variants with gradient recovery use 208 computed updates. The reconstruction ablation keeps the selected mask and directly slices the dense weights.

\subsection{Hyperparameter Sensitivity}
\label{app:hyperparameter_sensitivity}

Figure~\ref{fig:ablation_hyperparameters} varies one coefficient at a time. The damping sweep uses $0.1\%$, $1\%$, and $10\%$ of the retained Gram matrix's mean diagonal with the same mask. The $\beta_r$ and $\beta_h$ sweeps each use $0.1$, $0.25$, and $0.5$ with the same reconstructed initializer.
\FloatBarrier

\section{Pruning Results Across Ratios and Architectures}
\label{app:detailed_results}

Tables~\ref{tab:main_results}--\ref{tab:additional_quality_surjo} extend Table~\ref{tab:summary_results} to additional pruning ratios and model configurations. Structured comparisons match parameter budgets and native loop counts, and sparse comparisons use the same target matrices and nonzero constraints. Evaluation follows Appendix~\ref{sec:Details_of_Benchmarks}.

\paragraph{Structured pruning at different ratios.}
In Table~\ref{tab:main_results}, B/16 averages 72.17 and 70.65 at 10\% and 20\% structured pruning, respectively. These scores exceed EcoDiff and PPCL by 0.85\% and 1.80\% in absolute terms, respectively. At 10\%, \method{} retains 99.79\% of the dense average.

\paragraph{Unstructured pruning at different ratios.}
Table~\ref{tab:unstructured_results} shows that B/16 averages 71.94 and 67.47 at 40\% and 60\% unstructured pruning, respectively, improving over 2ndMatch by 0.69\% and 3.13\%. At 50\% on B/32, \method{} raises the average from 65.86 to 65.90.

\paragraph{2:4 semi-structured pruning.}
On B/32, \method{} leads the pruning methods on all six benchmarks in Table~\ref{tab:semi_structured_results}. Its average is 63.25, an absolute gain of 0.51\% over 2ndMatch.

\begin{table*}[!tp]
\centering
\captionsetup{skip=2pt}
\footnotesize
\setlength{\tabcolsep}{1.5pt}
\renewcommand{\arraystretch}{0.96}
\caption{\textbf{Looped-DiT B/16 at 10\% and 20\% structured parameter pruning.} Average is the unweighted mean of six benchmarks; bold marks the best pruning score within a configuration.}
\label{tab:main_results}
\label{tab:different_budgets_results}
\label{tab:appendix_structured_results}
\begin{tabularx}{\textwidth}{@{}>{\raggedright\arraybackslash}p{0.33\textwidth}*{7}{>{\centering\arraybackslash}X}@{}}
\toprule
& \multicolumn{3}{c}{\textbf{Alignment}} & \multicolumn{3}{c}{\textbf{Reasoning and spatial}} & \textbf{Overall} \\
\cmidrule(lr){2-4}\cmidrule(lr){5-7}\cmidrule(l){8-8}
\textbf{Method} & \textbf{GenEval} & \textbf{DPG} & \textbf{TIIF} & \textbf{PRISM} & \textbf{CoRe} & \textbf{Spatial} & \textbf{Average} \\
\midrule
\multicolumn{8}{c}{\cellcolor{kvsoftheader}\textbf{Looped-DiT B/16 ($n_{\ell}=4$) --- 10\% parameter pruning}}\\
\midrule
\rowcolor{kvsoftgray}Dense & 86.06 & 86.98 & 76.14 & 66.49 & 53.67 & 64.61 & 72.32 \\
Wanda-sp~(ICLR'24) & 78.89 & 85.19 & 72.44 & 58.86 & 49.50 & 63.45 & 68.05 \\
SlimGPT~(NeurIPS'24) & 83.18 & 85.35 & 72.55 & 57.99 & 49.88 & 64.42 & 68.90 \\
MosaicDiff~(ICCV'25) & 82.06 & 85.40 & 72.09 & 59.46 & 50.09 & 64.04 & 68.86 \\
OBS-Diff~(ICLR'26) & 83.75 & 86.03 & 73.26 & 58.09 & 51.25 & 64.49 & 69.48 \\
Importance-aware OBS~(NeurIPS'26) & 83.44 & 84.79 & 73.38 & 57.02 & 51.95 & 64.98 & 69.26 \\
EcoDiff~(ICLR'26) & \textbf{86.86} & 86.28 & 75.42 & 60.69 & 53.24 & 65.43 & 71.32 \\
PPCL~(CVPR'26) & 86.08 & 86.52 & 74.13 & 61.46 & 51.85 & 64.79 & 70.81 \\
MobileWan~(NeurIPS'26) & 84.69 & 86.44 & 75.13 & 60.64 & 52.36 & \textbf{65.55} & 70.80 \\
\rowcolor{kvsoftcyan}\textbf{\method{}} & 86.21 & \textbf{87.40} & \textbf{76.13} & \textbf{62.41} & \textbf{55.34} & 65.52 & \textbf{72.17} \\
\midrule
\multicolumn{8}{c}{\cellcolor{kvsoftheader}\textbf{Looped-DiT B/16 ($n_{\ell}=4$) --- 20\% parameter pruning}}\\
\midrule
Wanda-sp~(ICLR'24) & 77.86 & 81.39 & 70.12 & 47.83 & 48.33 & 63.52 & 64.84 \\
SlimGPT~(NeurIPS'24) & 76.22 & 81.92 & 68.65 & 49.14 & 44.07 & 62.07 & 63.68 \\
MosaicDiff~(ICCV'25) & 62.91 & 75.91 & 59.22 & 44.66 & 41.37 & 57.11 & 56.86 \\
OBS-Diff~(ICLR'26) & 74.33 & 79.84 & 65.75 & 47.79 & 41.99 & 59.76 & 61.57 \\
Importance-aware OBS~(NeurIPS'26) & 75.08 & 78.21 & 66.06 & 47.71 & 40.63 & 60.05 & 61.29 \\
EcoDiff~(ICLR'26) & 77.37 & 84.86 & 72.61 & 51.78 & 51.50 & 64.32 & 67.07 \\
PPCL~(CVPR'26) & 83.12 & 85.55 & 74.04 & 54.48 & 50.59 & \textbf{65.33} & 68.85 \\
MobileWan~(NeurIPS'26) & 76.24 & 84.50 & 69.94 & 51.08 & 46.27 & 64.04 & 65.34 \\
\rowcolor{kvsoftcyan}\textbf{\method{}} & \textbf{87.51} & \textbf{85.74} & \textbf{74.11} & \textbf{58.44} & \textbf{52.87} & 65.21 & \textbf{70.65} \\
\bottomrule
\end{tabularx}
\end{table*}

\begin{table*}[!tp]
\centering
\captionsetup{skip=2pt}
\footnotesize
\setlength{\tabcolsep}{1.5pt}
\renewcommand{\arraystretch}{0.96}
\caption{\textbf{Unstructured pruning results for Looped-DiT B/16 and B/32.} Average is the unweighted mean of six benchmarks; bold marks the best pruning score within a configuration.}
\label{tab:unstructured_results}
\label{tab:unstructured_multimodel_results}
\label{tab:appendix_unstructured_results}
\begin{tabular}{@{}>{\raggedright\arraybackslash}p{.33\textwidth}*{7}{>{\centering\arraybackslash}p{\dimexpr(\textwidth-.33\textwidth)/7-2\tabcolsep\relax}}@{}}
\toprule
& \multicolumn{3}{c}{\textbf{Alignment}} & \multicolumn{3}{c}{\textbf{Reasoning and spatial}} & \textbf{Overall} \\
\cmidrule(lr){2-4}\cmidrule(lr){5-7}\cmidrule(l){8-8}
\textbf{Method} & \textbf{GenEval} & \textbf{DPG} & \textbf{TIIF} & \textbf{PRISM} & \textbf{CoRe} & \textbf{Spatial} & \textbf{Average} \\
\midrule
\multicolumn{8}{c}{\cellcolor{kvsoftheader}\textbf{Looped-DiT B/16 ($n_{\ell}=4$) --- 40\% weight sparsity}}\\
\rowcolor{kvsoftgray}Dense & 86.06 & 86.98 & 76.14 & 66.49 & 53.67 & 64.61 & 72.32 \\
Wanda~(ICLR'24) & 84.52 & 86.21 & 74.26 & 61.06 & 52.91 & 64.19 & 70.53 \\
SparseGPT~(ICML'23) & 83.06 & 85.96 & 74.38 & 62.84 & 52.04 & 65.02 & 70.55 \\
OBS-Diff~(ICLR'26) & 83.18 & 85.68 & 74.13 & 63.04 & 52.30 & 64.98 & 70.55 \\
Importance-aware OBS~(NeurIPS'26) & 83.46 & 85.92 & 74.73 & 62.89 & 51.87 & 65.28 & 70.69 \\
2ndMatch~(CVPR'26) & 84.33 & 86.54 & 75.26 & 64.02 & 52.14 & 65.19 & 71.25 \\
\rowcolor{kvsoftcyan}\textbf{\method{}} & \textbf{86.77} & \textbf{86.71} & \textbf{75.67} & \textbf{64.10} & \textbf{52.92} & \textbf{65.46} & \textbf{71.94} \\
\midrule
\multicolumn{8}{c}{\cellcolor{kvsoftheader}\textbf{Looped-DiT B/16 ($n_{\ell}=4$) --- 60\% weight sparsity}}\\
Wanda~(ICLR'24) & 13.29 & 53.31 & 42.40 & 18.76 & 26.57 & 43.89 & 33.04 \\
SparseGPT~(ICML'23) & 44.39 & 68.33 & 56.85 & 33.19 & 30.00 & 53.52 & 47.71 \\
OBS-Diff~(ICLR'26) & 48.26 & 69.75 & 58.00 & 34.92 & 30.69 & 54.03 & 49.27 \\
Importance-aware OBS~(NeurIPS'26) & 56.54 & 72.93 & 62.03 & 35.64 & 32.19 & 58.18 & 52.92 \\
2ndMatch~(CVPR'26) & 78.74 & 82.23 & 68.76 & 47.81 & 45.16 & 63.35 & 64.34 \\
\rowcolor{kvsoftcyan}\textbf{\method{}} & \textbf{82.71} & \textbf{83.71} & \textbf{72.50} & \textbf{52.46} & \textbf{48.98} & \textbf{64.49} & \textbf{67.47} \\
\midrule
\multicolumn{8}{c}{\cellcolor{kvsoftheader}\textbf{Looped-DiT B/32 ($n_{\ell}=4$) --- 50\% weight sparsity}}\\
\rowcolor{kvsoftgray}Dense & 84.49 & 85.21 & 71.95 & 53.53 & 44.79 & 63.80 & 67.30 \\
Wanda~(ICLR'24) & 79.19 & 82.77 & 69.78 & 49.98 & 42.94 & 62.86 & 64.59 \\
SparseGPT~(ICML'23) & 79.59 & 83.00 & 69.98 & 49.33 & 43.06 & 63.77 & 64.79 \\
OBS-Diff~(ICLR'26) & 79.49 & 83.02 & 69.23 & 49.79 & 42.95 & 63.59 & 64.68 \\
Importance-aware OBS~(NeurIPS'26) & 78.74 & 82.63 & 70.19 & 49.69 & 42.68 & 63.19 & 64.52 \\
2ndMatch~(CVPR'26) & \textbf{81.92} & 83.81 & 70.92 & \textbf{50.11} & 44.34 & \textbf{64.07} & 65.86 \\
\rowcolor{kvsoftcyan}\textbf{\method{}} & 81.42 & \textbf{83.98} & \textbf{70.97} & 50.10 & \textbf{44.88} & \textbf{64.07} & \textbf{65.90} \\
\bottomrule
\end{tabular}
\end{table*}

\begin{table*}[!tp]
\centering
\captionsetup{skip=2pt}
\footnotesize
\setlength{\tabcolsep}{1.5pt}
\renewcommand{\arraystretch}{0.85}
\caption{\textbf{Looped-DiT B/32 at 2:4 semi-structured pruning.} Average is the unweighted mean of six benchmarks; bold marks the best pruning score within a configuration.}
\label{tab:semi_structured_results}
\label{tab:appendix_2of4_results}
\begin{tabularx}{\textwidth}{@{}>{\raggedright\arraybackslash}p{0.33\textwidth}*{7}{>{\centering\arraybackslash}X}@{}}
\toprule
& \multicolumn{3}{c}{\textbf{Alignment}} & \multicolumn{3}{c}{\textbf{Reasoning and spatial}} & \textbf{Overall} \\
\cmidrule(lr){2-4}\cmidrule(lr){5-7}\cmidrule(l){8-8}
\textbf{Method} & \textbf{GenEval} & \textbf{DPG} & \textbf{TIIF} & \textbf{PRISM} & \textbf{CoRe} & \textbf{Spatial} & \textbf{Average} \\
\midrule
\multicolumn{8}{c}{\cellcolor{kvsoftheader}\textbf{Looped-DiT B/32 ($n_{\ell}=4$) --- 2:4 weight sparsity}}\\
\midrule
\rowcolor{kvsoftgray}Dense & 84.49 & 85.21 & 71.95 & 53.53 & 44.79 & 63.80 & 67.30 \\
Wanda~(ICLR'24) & 58.74 & 74.03 & 64.63 & 35.29 & 35.08 & 59.88 & 54.61 \\
SparseGPT~(ICML'23) & 73.53 & 80.07 & 66.96 & 42.46 & 39.04 & 62.32 & 60.73 \\
OBS-Diff~(ICLR'26) & 73.63 & 79.53 & 67.38 & 42.83 & 39.64 & 61.82 & 60.81 \\
Importance-aware OBS~(NeurIPS'26) & 73.23 & 78.55 & 66.84 & 43.60 & 38.77 & 61.76 & 60.46 \\
2ndMatch~(CVPR'26) & 77.76 & 81.46 & 68.72 & 44.61 & 41.45 & 62.46 & 62.74 \\
\rowcolor{kvsoftcyan}\textbf{\method{}} & \textbf{79.44} & \textbf{81.70} & \textbf{69.26} & \textbf{44.92} & \textbf{41.58} & \textbf{62.59} & \textbf{63.25} \\
\bottomrule
\end{tabularx}
\end{table*}

\subsection{Results on Additional Model Configurations}
\label{app:sdxsv3_additional_results}

\paragraph{SDXSv3.}
Table~\ref{tab:sdxsv3_additional_quality} covers the three configurations timed in Figure~\ref{fig:efficiency_results}. At 30\% structured, 2:4 semi-structured, and 60\% unstructured pruning, \method{} averages 42.89, 42.09, and 42.56. The respective gains are 0.39\%, 0.49\%, and 0.22\% over OBS-Diff in the first setting and 2ndMatch in the other two.

\paragraph{Looped-DiT B/32.}
In Table~\ref{tab:additional_quality_b32}, \method{} averages 64.87 at 30\% structured pruning and 63.49 at 60\% unstructured pruning. These scores exceed OBS-Diff and 2ndMatch by 3.27\% and 0.43\%, respectively. At 30\%, it leads pruning methods on all six benchmarks.

\paragraph{Surjo-Image-Preview.}
In Table~\ref{tab:additional_quality_surjo}, \method{} averages 40.11 at 30\% structured pruning and 40.41 at 60\% unstructured pruning, exceeding EcoDiff and Importance-aware OBS by 0.71\% and 1.28\%, respectively. At 60\%, it leads pruning methods on all six benchmarks.

\begin{table*}[!tp]
\centering
\footnotesize
\setlength{\tabcolsep}{3pt}
\renewcommand{\arraystretch}{0.72}
\caption{\textbf{SDXSv3 results across three pruning formats.} Average is the unweighted mean of six benchmarks; bold marks the best pruning score within a configuration.}
\label{tab:sdxsv3_additional_quality}
\begin{tabular}{@{}>{\raggedright\arraybackslash}p{.30\textwidth}*{7}{>{\centering\arraybackslash}p{\dimexpr(\textwidth-.30\textwidth)/7-2\tabcolsep\relax}}@{}}
\toprule
\textbf{Method} & \textbf{GenEval} & \textbf{DPG} & \textbf{TIIF} & \textbf{PRISM} & \textbf{CoRe} & \textbf{Spatial} & \textbf{Average} \\
\midrule
\rowcolor{kvsoftgray}Dense & 26.35 & 63.57 & 50.09 & 40.36 & 29.98 & 51.71 & 43.67 \\
\midrule
\multicolumn{8}{c}{\cellcolor{kvsoftheader}\textbf{30\% structured pruning}} \\
Wanda-sp & 23.02 & 64.45 & 47.06 & 34.71 & 29.19 & 51.33 & 41.63 \\
SlimGPT & 24.08 & 65.54 & 48.19 & 35.90 & 29.47 & 51.45 & 42.44 \\
MosaicDiff & 22.91 & 60.81 & 45.07 & 32.92 & 27.32 & 48.23 & 39.54 \\
OBS-Diff & 24.00 & \textbf{65.66} & 47.71 & 36.21 & 29.97 & 51.45 & 42.50 \\
Importance-aware OBS & 23.13 & 63.10 & 46.91 & 35.40 & 28.79 & 51.27 & 41.44 \\
EcoDiff & 22.93 & 60.34 & 48.67 & 34.46 & 29.85 & 50.50 & 41.12 \\
PPCL & 24.63 & 60.21 & 47.76 & 37.22 & 28.84 & 50.60 & 41.54 \\
MobileWan & 23.11 & 61.28 & 47.58 & 37.44 & 28.15 & 49.57 & 41.19 \\
\rowcolor{kvsoftcyan}\textbf{\method{}} & \textbf{25.01} & 63.50 & \textbf{49.41} & \textbf{37.50} & \textbf{30.00} & \textbf{51.91} & \textbf{42.89} \\
\midrule
\multicolumn{8}{c}{\cellcolor{kvsoftheader}\textbf{2:4 semi-structured pruning}} \\
Wanda & 8.14 & 38.99 & 36.80 & 21.66 & 13.95 & 24.05 & 23.93 \\
SparseGPT & 20.98 & 59.24 & 47.36 & 35.63 & 26.87 & 49.21 & 39.88 \\
OBS-Diff & 21.29 & 59.71 & 48.02 & 35.34 & 26.47 & 49.54 & 40.06 \\
Importance-aware OBS & 22.22 & 58.62 & 47.81 & 35.34 & 26.40 & 48.99 & 39.90 \\
2ndMatch & 24.16 & 60.68 & 48.65 & 35.33 & \textbf{29.44} & \textbf{51.33} & 41.60 \\
\rowcolor{kvsoftcyan}\textbf{\method{}} & \textbf{24.40} & \textbf{61.21} & \textbf{49.21} & \textbf{37.48} & 28.89 & \textbf{51.33} & \textbf{42.09} \\
\midrule
\multicolumn{8}{c}{\cellcolor{kvsoftheader}\textbf{60\% unstructured pruning}} \\
Wanda & 3.02 & 34.04 & 30.89 & 18.49 & 10.70 & 23.60 & 20.12 \\
SparseGPT & 21.38 & 58.32 & 47.08 & 34.68 & 24.79 & 48.86 & 39.18 \\
OBS-Diff & 21.88 & 58.53 & 47.37 & 35.20 & 25.33 & 49.67 & 39.67 \\
Importance-aware OBS & 21.41 & 58.29 & 47.72 & 34.21 & 25.53 & 49.20 & 39.39 \\
2ndMatch & 23.61 & 62.71 & 49.07 & 36.06 & \textbf{29.72} & \textbf{52.86} & 42.34 \\
\rowcolor{kvsoftcyan}\textbf{\method{}} & \textbf{23.99} & \textbf{62.99} & \textbf{49.45} & \textbf{37.51} & 28.91 & 52.53 & \textbf{42.56} \\
\bottomrule
\end{tabular}
\end{table*}

\begin{table*}[!tp]
\centering
\footnotesize
\setlength{\tabcolsep}{3pt}
\caption{\textbf{Looped-DiT B/32 at 30\% structured and 60\% unstructured pruning.} Average is the unweighted mean of six benchmarks; bold marks the best pruning score within a configuration.}
\label{tab:additional_quality_b32}
\begin{tabular}{@{}>{\raggedright\arraybackslash}p{.30\textwidth}*{7}{>{\centering\arraybackslash}p{\dimexpr(\textwidth-.30\textwidth)/7-2\tabcolsep\relax}}@{}}
\toprule
\textbf{Method} & \textbf{GenEval} & \textbf{DPG} & \textbf{TIIF} & \textbf{PRISM} & \textbf{CoRe} & \textbf{Spatial} & \textbf{Average} \\
\midrule
\rowcolor{kvsoftgray}Dense & 84.49 & 85.21 & 71.95 & 53.53 & 44.79 & 63.80 & 67.30 \\
\midrule
\multicolumn{8}{c}{\cellcolor{kvsoftheader}\textbf{30\% structured pruning}} \\
Wanda-sp & 7.66 & 60.61 & 34.42 & 20.03 & 24.31 & 50.91 & 32.99 \\
SlimGPT & 76.08 & 81.16 & 67.80 & 42.38 & 39.14 & 62.46 & 61.50 \\
MosaicDiff & 61.87 & 78.58 & 63.95 & 37.62 & 38.73 & 60.67 & 56.90 \\
OBS-Diff & 75.93 & 80.32 & 67.76 & 44.47 & 38.86 & 62.24 & 61.60 \\
Importance-aware OBS & 74.95 & 80.21 & 67.25 & 42.44 & 38.69 & 62.54 & 61.01 \\
EcoDiff & 70.32 & 80.64 & 68.14 & 38.64 & 42.00 & 62.28 & 60.34 \\
PPCL & 61.47 & 75.87 & 64.46 & 35.96 & 36.89 & 58.79 & 55.57 \\
MobileWan & 45.85 & 75.78 & 61.36 & 33.11 & 34.19 & 57.91 & 51.37 \\
\rowcolor{kvsoftcyan}\textbf{\method{}} & \textbf{82.70} & \textbf{82.04} & \textbf{70.17} & \textbf{47.84} & \textbf{43.41} & \textbf{63.07} & \textbf{64.87} \\
\midrule
\multicolumn{8}{c}{\cellcolor{kvsoftheader}\textbf{60\% unstructured pruning}} \\
Wanda & 51.05 & 76.55 & 60.44 & 37.76 & 32.67 & 59.33 & 52.96 \\
SparseGPT & 71.88 & 79.24 & 66.77 & 42.36 & 40.05 & 62.46 & 60.46 \\
OBS-Diff & 71.82 & 79.74 & 67.40 & 42.02 & 39.78 & 61.51 & 60.38 \\
Importance-aware OBS & 71.13 & 79.71 & 67.52 & 41.69 & 38.38 & 61.20 & 59.94 \\
2ndMatch & 78.02 & 81.74 & 69.10 & 44.44 & \textbf{42.34} & 62.69 & 63.06 \\
\rowcolor{kvsoftcyan}\textbf{\method{}} & \textbf{79.23} & \textbf{82.03} & \textbf{69.56} & \textbf{44.79} & 42.31 & \textbf{63.01} & \textbf{63.49} \\
\bottomrule
\end{tabular}
\end{table*}

\begin{table*}[!tp]
\centering
\footnotesize
\setlength{\tabcolsep}{3pt}
\caption{\textbf{Surjo-Image-Preview at 30\% structured and 60\% unstructured pruning.} Average is the unweighted mean of six benchmarks; bold marks the best pruning score within a configuration.}
\label{tab:additional_quality_surjo}
\begin{tabular}{@{}>{\raggedright\arraybackslash}p{.30\textwidth}*{7}{>{\centering\arraybackslash}p{\dimexpr(\textwidth-.30\textwidth)/7-2\tabcolsep\relax}}@{}}
\toprule
\textbf{Method} & \textbf{GenEval} & \textbf{DPG} & \textbf{TIIF} & \textbf{PRISM} & \textbf{CoRe} & \textbf{Spatial} & \textbf{Average} \\
\midrule
\rowcolor{kvsoftgray}Dense & 32.92 & 73.77 & 52.33 & 46.09 & 34.23 & 52.93 & 48.71 \\
\midrule
\multicolumn{8}{c}{\cellcolor{kvsoftheader}\textbf{30\% structured pruning}} \\
Wanda-sp & 0.83 & 18.20 & 22.09 & 7.43 & 3.99 & 5.22 & 9.63 \\
SlimGPT & 13.58 & 56.50 & 40.24 & 21.99 & 19.63 & 38.20 & 31.69 \\
MosaicDiff & 12.86 & 56.19 & 39.23 & 24.54 & 19.05 & 40.57 & 32.07 \\
OBS-Diff & 13.19 & 55.20 & 39.00 & 22.05 & 17.80 & 38.59 & 30.97 \\
Importance-aware OBS & 15.22 & 58.18 & 40.96 & 23.35 & 19.94 & 42.37 & 33.34 \\
EcoDiff & 19.00 & \textbf{65.99} & 46.89 & 30.60 & 26.56 & 47.37 & 39.40 \\
PPCL & 19.08 & 63.03 & 46.74 & 30.07 & 24.78 & 46.34 & 38.34 \\
MobileWan & 13.52 & 54.54 & 39.93 & 24.61 & 22.98 & 35.92 & 31.92 \\
\rowcolor{kvsoftcyan}\textbf{\method{}} & \textbf{20.58} & 65.96 & \textbf{47.94} & \textbf{30.81} & \textbf{27.88} & \textbf{47.51} & \textbf{40.11} \\
\midrule
\multicolumn{8}{c}{\cellcolor{kvsoftheader}\textbf{60\% unstructured pruning}} \\
Wanda & 13.40 & 55.37 & 42.57 & 27.88 & 20.92 & 40.45 & 33.43 \\
SparseGPT & 19.23 & 63.86 & 44.85 & 31.17 & 24.78 & 46.38 & 38.38 \\
OBS-Diff & 19.81 & 64.86 & 45.39 & 31.24 & 24.95 & 46.36 & 38.77 \\
Importance-aware OBS & 19.59 & 64.83 & 45.37 & 32.07 & 26.81 & 46.07 & 39.13 \\
2ndMatch & 18.18 & 59.82 & 42.03 & 27.27 & 23.03 & 43.33 & 35.61 \\
\rowcolor{kvsoftcyan}\textbf{\method{}} & \textbf{20.46} & \textbf{65.95} & \textbf{46.68} & \textbf{34.41} & \textbf{27.11} & \textbf{47.82} & \textbf{40.41} \\
\bottomrule
\end{tabular}
\end{table*}
\FloatBarrier
\clearpage
\begin{figure*}[p]
\section{Qualitative Evaluation}
\label{sec:more_qualitative}
\subsection{Generated Images of the Observation Controls}
\label{app:qualitative_controls}
\centering
\includegraphics[width=\appendixfigurewidth,height=.40\textheight,keepaspectratio]{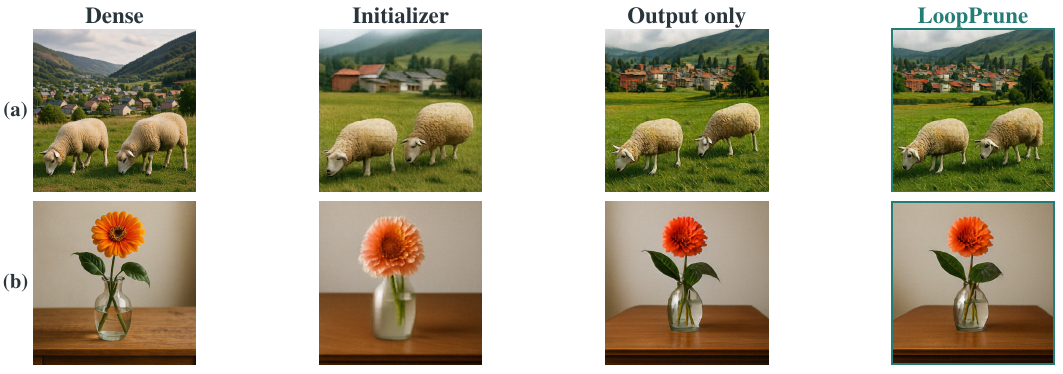}
\caption{\textbf{Recovery restores visual details after pruning under a matched update budget.} Two selected held-out prompts compare Dense, the pruned initializer, output-only recovery, and LoopPrune. Both recovery variants compute 208 updates, and all models use the same noise per prompt. LoopPrune retains two grazing sheep and the hillside village in (a), and shows distinct leaves and a clear vase in (b).}
\label{fig:observation_gallery_1}
\end{figure*}

\begin{figure*}[p]
\subsection{Comparison with Pruning Baselines}
\label{app:qualitative_current}
\centering
\includegraphics[width=\appendixfigurewidth,height=.40\textheight,keepaspectratio]{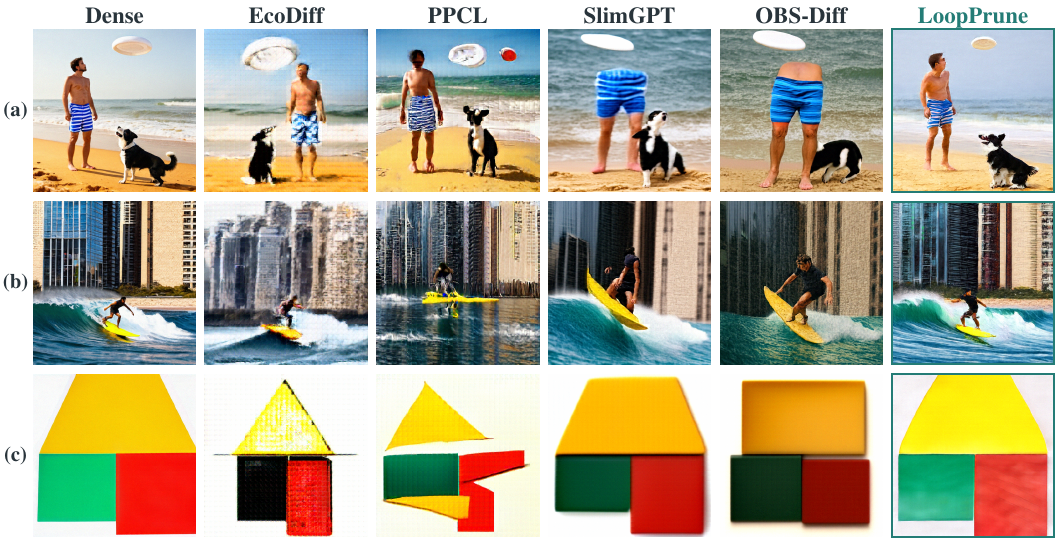}
\caption{\textbf{Qualitative comparison at 30\% structured pruning on Looped-DiT B/16.} Three selected DPG-Bench prompts describe (a) a beach frisbee scene, (b) a surfer against a city skyline, and (c) colored geometric shapes. For each prompt, all methods use the same initial noise and sampling settings. LoopPrune more closely preserves the subject layout and key details of the dense model in these examples.}
\label{fig:current_gallery_one}
\end{figure*}
\FloatBarrier

\end{document}